\documentclass[lettersize,journal]{IEEEtran}
\usepackage{amsmath,amsfonts,amssymb}
\usepackage{algorithmic}
\usepackage{array}
\usepackage[caption=false,font=normalsize,labelfont=sf,textfont=sf]{subfig}
\usepackage{textcomp}
\usepackage{stfloats}
\usepackage{url}
\usepackage{verbatim}
\usepackage{graphicx}
\usepackage{booktabs}
\usepackage{multirow}
\usepackage{threeparttable}
\def\BibTeX{{\rm B\kern-.05em{\sc i\kern-.025em b}\kern-.08em
    T\kern-.1667em\lower.7ex\hbox{E}\kern-.125emX}}
\usepackage{balance}
\begin{document}
\title{Latency Coding for Efficient and Low-Latency Deep Spiking Neural Networks}
\author{Yi Lu$^1$, Jianhao Ding$^2$, Zhaofei Yu$^3$
\thanks{$^1$ School of Software and Microelectronics, Peking University, Beijing, China}
\thanks{$^2$ School of Computer Science, Peking University, Beijing, China}
\thanks{$^3$ Institute for Artificial Intelligence and the School of Computer Science, Peking University, Beijing, China (e-mail: yuzf12@pku.edu.cn).}
\thanks{This work is supported by the National Natural Science Foundation of China (U24B20140,62422601), Beijing Municipal Science and Technology Program (Z241100004224004), Beijing Nova Program (20230484362, 20240484703), National Key Laboratory for Multimedia Information Processing, and Beijing Key Laboratory of Brain-inspired Spiking Large Models.}
}
\markboth{IEEE Transactions on Emerging Topics in Computational Intelligence}{Under Review}
\maketitle

\begin{abstract}
    Spiking neural networks (SNNs) offer a biologically inspired computing paradigm with significant potential for energy-efficient neural processing. Among neural coding schemes of SNNs, Time-To-First-Spike (TTFS) coding, which encodes information through the precise timing of a neuron's first spike, provides exceptional activity sparsity and energy efficiency. However, existing TTFS models lack efficient training methods, suffering from high inference latency and limited performance, limiting their practicality on neuromorphic hardware. In this work, we propose latency coding, an extension of TTFS coding, and present a compatible framework that enables the efficient training of deep latency-coded SNNs by leveraging backpropagation through time (BPTT) algorithm. The framework includes: (1) a latency encoding (LE) module with feature extraction and straight-through estimators to address severe information loss in direct intensity-to-latency mapping; (2) relaxation of the strict single-spike constraint in intermediate layers to improve information propagation and gradient flow; and (3) a temporal adaptive decision (TAD) loss function that dynamically weights supervision signals based on the model's confidence, balancing the trade-off between speed and accuracy. Experimental results demonstrate that our method achieves competitive or superior accuracy compared with existing TTFS-coded SNNs with ultra-low inference latency and high energy efficiency. Latency-coded SNNs also demonstrate improved robustness against input perturbations. These findings highlight latency coding as a practical and hardware-friendly approach for fast and energy-efficient neuromorphic processing.

\end{abstract}


\section{Introduction}
\IEEEPARstart{E}{merging} as an alternative paradigm for artificial intelligence, spiking neural networks (SNNs) have gained increasing attention due to their biological plausibility and superior energy efficiency. Inspired by the neuronal functionality of biological brains, SNNs encode information through binary spikes, exhibiting distinctive features of sparse and asynchronous processing, which endow them with rich spatial-temporal dynamics, exceptional information processing capability, and low power consumption. By leveraging these advantages, SNNs play a pivotal role in neuromorphic computing development, demonstrating significant potential to overcome the limitations of conventional artificial neural networks (ANNs) in low-power, real-time scenarios. Over the past decade, significant advancements in training algorithms and architectural designs have enhanced the capabilities of SNNs, enabling their successful deployment across various applications, including image classification \cite{ding2021optimal,qin2023attention,wu2021tandem}, autonomous vehicle control \cite{bing2018end,bing2020indirect}, auditory processing \cite{wu2021progressive,tang2024neuromorphic}, and edge computing \cite{liu2026fpf}, where their energy-efficient, event-driven computation offers distinct advantages.

Neural coding schemes are fundamental to information representation and transmission in SNNs. To date, a variety of neural coding schemes inspired by neuroscience have been applied in SNNs. Among these, rate coding stands out as the simplest and most widely adopted approach, encoding information in the frequency of spikes emitted by a neuron over a given period, where a higher firing rate signifies a stronger signal. However, it is unable to fully utilize the temporal information in spike trains, and falls short in energy efficiency due to the high spikes volume. In contrast, Time-To-First-Spike (TTFS) coding, a form of temporal coding, offers a highly energy-efficient way to convey information. It encodes signals into the precise firing time or latency of a neuron's spike, typically following the principle of ``larger values trigger earlier spikes''. This mechanism inherently optimizes for energy efficiency, making it an ideal strategy for low-power applications. 

However, despite these theoretical advantages, realizing efficient TTFS models in practice remains challenging. Existing TTFS approaches predominantly rely on event-based learning rules or ANN-to-SNN conversion, making it difficult to train deep TTFS-coded SNNs with competitive accuracy. Moreover, existing modeling approaches of TTFS neurons transmit precise firing times between layers, where these firing times are usually represented as floating-point values during the forward pass and training. Such representations are not natively supported by existing neuromorphic hardware, which is designed to communicate discrete binary spikes. To deploy these models on neuromorphic platforms, the floating-point firing times must be quantized, which inevitably downgrades performance. In addition, time-based inter-layer synchronization makes the total timesteps needed for inference scale with both the number of quantized steps and the number of layers. Therefore, traditional TTFS methods usually require hundreds or even thousands of timesteps.

In parallel with the development of temporal coding, recent years have witnessed the significant progress of rate-coded SNNs applying backpropagation through time (BPTT) with surrogate gradients. Surprisingly, despite the success of BPTT in rate coding, its application to TTFS coding remains largely unexplored. This motivates us to explore a practical and energy-efficient coding scheme that combines the sparse temporal activity of TTFS with the trainability of BPTT. By integrating BPTT into TTFS-style representations, our framework aims to preserve the low-energy advantage of temporal coding while sharply reducing the inference latency of deep SNNs.

In this work, we present a new framework to incorporate the BPTT training paradigm into TTFS-coded SNNs. This integration, however, must overcome three primary obstacles. First, directly encoding input images into temporal spike trains leads to severe information loss. For this, we design an encoding module that extracts initial features of inputs and subsequently encodes them into spikes. Second, the strict single-spike constraint of conventional TTFS coding incurs information decay and gradient diminishing as the network becomes deeper. To tackle this issue, our approach strategically allows neurons in intermediate layers to fire multiple times, while still using the first spike of the output layer as the decision signal. To distinguish it from existing single-spike TTFS methods, we refer to this generalized TTFS formulation as latency coding hereafter. The last obstacle lies in the incompatibility of latency coding with standard cross-entropy loss. Standard cross-entropy loss treats each timestep equally, which expects the network to make a decision based on the spike rate over all timesteps, thus conflicting with the temporal nature of latency coding. To resolve this, we introduce a loss function that dynamically weights supervision signals to balance inference speed and accuracy.

Beyond addressing these training obstacles, latency coding offers key advantages over conventional TTFS. It preserves the first-spike decision principle at the output layer, but relaxes the strict single-spike constraint in intermediate layers to improve information propagation and gradient flow. Unlike existing TTFS methods that transmit floating-point firing times between layers, latency coding propagates binary spikes through the network, which is natively compatible with neuromorphic hardware designed for discrete spike communication. This makes latency coding practically deployable on existing neuromorphic platforms without the performance degradation caused by quantization. Furthermore, by enabling end-to-end BPTT training without relying on inter-layer time synchronization, our framework drastically reduces the number of required inference timesteps compared to conventional TTFS approaches, closing the latency gap with rate-coded SNNs while retaining their energy efficiency advantage. Our main contributions can be summarized as follows:
\begin{itemize}
    \item {We propose latency coding as an extension of TTFS coding. By preserving the first-spike decision principle while using binary spike communication between layers, latency coding maintains the energy-efficient nature of temporal coding, substantially reduces inference latency, and enables direct deployment on existing neuromorphic hardware.}
    \item {We develop a practical training framework that integrates the BPTT paradigm with latency-coded SNNs. By relaxing the strict single-spike constraint in intermediate layers, the framework provides richer temporal gradient paths for deep networks, while the membrane-potential-assisted decoding mechanism ensures deterministic first-spike decision-making at the output layer.}
    \item {We design two key components to address the main optimization obstacles of latency-coded SNNs. The latency encoding (LE) module reduces input information loss caused by intensity-to-latency mapping through feature extraction and straight-through estimation. The temporal adaptive decision (TAD) loss function resolves the incompatibility between latency coding and standard cross-entropy loss, dynamically weighting supervision signals based on the model's confidence to balance the trade-off between speed and accuracy.}
    \item {Extensive experiments on CIFAR-10, CIFAR-100, Tiny-ImageNet, ImageNet and neuromorphic datasets demonstrate that our method outperforms existing TTFS coding methods by achieving superior accuracy with ultra-low inference latency while maintaining energy efficiency. Furthermore, compared to rate-coded SNNs, our framework not only attains comparable classification performance, but also exhibits enhanced robustness against input perturbations.}
\end{itemize}

\section{Background and Related Works}
\subsection{Spiking Neuron Model}
\subsubsection{Spiking Neuron} 
The fundamental processing unit of an SNN is the spiking neuron model, which governs how incoming spikes are integrated and when an output spike is generated. A spectrum of neuron models exists, ranging from the biologically detailed and computationally expensive Hodgkin-Huxley model \cite{1952A} to more abstract and efficient models.

In this work, we employ the Leaky Integrate-and-Fire (LIF) model, which is one of the most widely used models in SNNs research. The LIF model offers an excellent trade-off between biological plausibility and computational efficiency, making it highly suitable for large-scale network simulations and implementations in neuromorphic hardware \cite{1997Networks}. The model captures two essential properties of a biological neuron: the integration of synaptic inputs over time and the passive decay (or ``leak'') of membrane potentials.
The subthreshold dynamics of the neuron's membrane potential, denoted by $u(t)$, can be described by the following first-order differential equation:
\begin{equation}
    \tau \frac{d u(t)}{d t}=-u(t)+I(t),
\end{equation}
where $u(t)$ represents the membrane potential at time $t$; $\tau$ is the membrane time constant, which governs the rate at which the potential decays back to its resting state; $I(t)$ is the total input current at time $t$, which is aggregated from the weighted spikes of all presynaptic neurons. The term $-u(t)$ represents the ``leaky'' property, which ensures that without sufficient input, the neuron's potential gradually returns to resting potential $0$. The ``integrate'' property is captured by the term $I(t)$, which drives the potential upwards in response to incoming stimuli.
\subsubsection{Iterative LIF Model}
For implementation in a digital simulation environment, the continuous dynamics are discretized into discrete timesteps of size $\Delta t$. By applying the exponential Euler method to the differential equation, we can derive an efficient iterative expression \cite{wu2019direct}. Mathematically, the subthreshold membrane potentials of neurons within an LIF layer are updated as:
\begin{equation}
    \boldsymbol{U}^l[t]=\tau \boldsymbol{U}^l[t-1] + W^l\boldsymbol{S}^{l-1}[t],
    \label{eq:lif_membrane}
\end{equation}
where $\tau$ is the constant leaky factor, $\boldsymbol{U}^l[t]$ denotes the membrane potentials of $l$-th layer at time $t$, and $W^l\boldsymbol{S}^{l-1}[t]$ represents the presynaptic inputs, which is the product of synaptic weights of the current layer $W^l$ and input spikes of the previous layer $\boldsymbol{S}^{l-1}[t]$.

The LIF neuron generates an output spike when its membrane potential reaches a predefined firing threshold, $V_{th}$. Once a spike is emitted, the membrane potential will be reset. Unlike the ``hard reset'' mechanism where the potential is clamped to a fixed value (e.g., the resting potential), we employ a ``soft reset'' mechanism. In this scheme, upon firing, the membrane potential is reduced by the value of the threshold $V_{th}$. This subtractive reset allows the neuron to retain a memory of any supra-threshold potential, which can influence the timing of subsequent spikes and is a feature observed in some biological neurons. Overall, the firing function and soft reset mechanism can be formulated as:
\begin{align}
\boldsymbol{S}^l[t]&=\Theta(\boldsymbol{U}^l[t]-V_{th}), \label{eq:lif_spike}\\
\boldsymbol{U}^l[t]&=\boldsymbol{U}^l[t] - \boldsymbol{S}^l[t] \cdot V_{th}, \label{eq:lif_reset}
\end{align}
where $\Theta(\cdot)$ denotes the Heaviside step function. The output spikes $\boldsymbol{S}^l[t]$ propagate to the subsequent layer, serving as the presynaptic input for its neurons.

\subsection{Training methods of SNNs}

ANN-to-SNN conversion is one of the earliest and most influential paradigms for obtaining high-performance spiking neural networks. The core idea is to first train an artificial neural network using standard backpropagation and then map the trained weights to an equivalent SNN, where the ReLU activation values are represented by spike firing rates over a temporal window. Early explorations of this idea can be traced back to \cite{perez2013mapping}, who investigated mappings from frame-driven to event-driven vision systems via rate coding. More recently, \cite{han2020rmp} identified the residual membrane potential as a key source of conversion error and proposed RMP-SNN, which calibrates the firing threshold to account for residual potentials, enabling deeper and more accurate converted SNNs with reduced timesteps. \cite{bu2023optimal} formulated the conversion problem from an optimization perspective and derived theoretically optimal conversion rules, which significantly narrows the efficiency gap between converted SNNs and their ANN counterparts. 

An alternative and increasingly popular paradigm is to directly train SNNs using gradient-based optimization. The central challenge of direct training lies in the non-differentiability of the spiking neuron's firing function. To overcome this, the surrogate gradient method has emerged as a practical solution. The key idea is to replace the gradient of the spike function with a smooth, continuous surrogate during the backward pass, while keeping the forward pass unchanged. \cite{neftci2019surrogate} provided a comprehensive review and theoretical justification of surrogate gradient learning. Building upon the surrogate gradient framework, backpropagation through time was introduced to handle the inherently temporal dynamics of SNNs. \cite{wu2018spatio} proposed spatio-temporal backpropagation, which unrolls the SNN across both spatial (layer) and temporal (timestep) dimensions and applies BPTT with surrogate gradients to jointly optimize the network over the entire spatio-temporal computational graph. \cite{ding2025assisting} explored the weight initialization method when using BPTT. Recent works further improve direct training through efficient spiking patterns and temporal-credit-assignment objectives \cite{shen2025exploiting,jiang2024deep}. Collectively, the surrogate gradient and BPTT framework have developed into a powerful and flexible paradigm for SNN training, progressively closing the accuracy gap with ANNs while preserving the energy-efficiency nature of spike-based computation.

\subsection{TTFS Coding}

In the context of spiking neural networks, time-to-first-spike as a coding scheme can be traced back to thesis by S. Thorpe \cite{thorpe1990spike}. Thorpe argues that the brain lacks sufficient time to process more than one spike from each neuron during a single processing procedure. Consequently, the first spike is expected to contain most of the essential information.  Analyses based on information-theoretic measures on experimental data have shown that the majority of the information about a new stimulus is indeed transmitted within the first 20 to 50 ms following the onset of the neuronal response \cite{optican1987temporal,tovee1993information,kjaer1994decoding,tovee1995information}, which validates Thorpe's claim. 

Building on this biological foundation, TTFS implements a temporal encoding strategy where the magnitude of an input signal is inversely mapped to the spike latency. In this scheme, a neuron receiving a stronger stimulus reaches its firing threshold more rapidly, thereby emitting a spike earlier; conversely, weaker stimuli result in delayed firing. Crucially, only the precise timing of this initial spike carries information, and any subsequent activity is typically ignored. This mechanism not only preserves essential analog information in the temporal domain but also ensures high computational efficiency and sparsity by limiting network activity to a single spike per neuron per inference cycle.

Training TTFS-based SNNs presents unique challenges due to its event-driven nature. Some works employed biologically plausible learning rules, such as Spike-Timing-Dependent Plasticity (STDP) \cite{kheradpisheh2018stdp} and Precise-Spike-Driven (PSD) Synaptic Plasticity \cite{yu2013precise}, to extract spatiotemporal features. However, scaling these methods to deep architectures proved difficult. Recently, supervised learning approaches based on exact spike time gradients have gained traction. \cite{mostafa2017supervised} demonstrated that by treating spike times as continuous variables, error gradients can be backpropagated directly. To handling the gradient explosion problem, \cite{zhang2021rectified} proposed a rectified linear postsynaptic potential function for spiking neurons. Based on this, \cite{wei2023temporal} introduced a novel training algorithm that enabling the efficient training of deep TTFS-coded SNNs.

\section{Methodology}

\begin{figure*}
    \centering
    \includegraphics[width=1\linewidth]{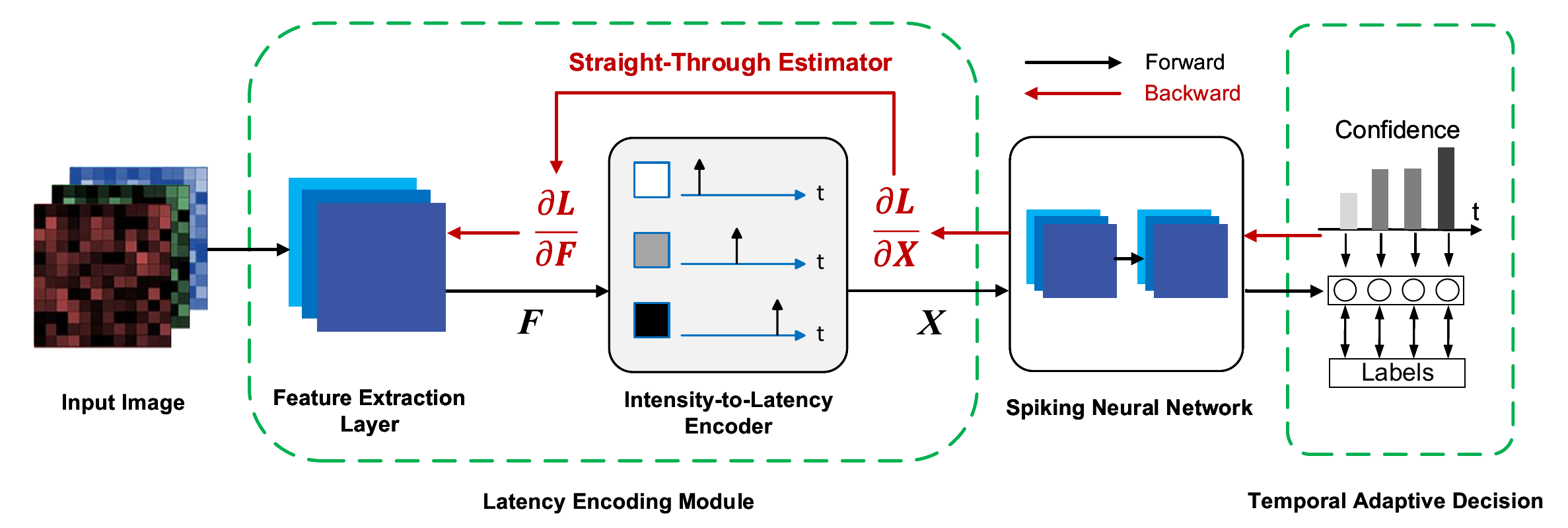}
    \caption{Overall training framework for the latency coding. The raw pixels of input images are first processed by the latency encoding module to generate temporal spike trains, which are then propagated through subsequent SNN layers. The TAD loss provides temporally adaptive supervision during training.}
    \label{framework}
\end{figure*}

\subsection{Latency Encoding Module}
As discussed earlier, it's improper for latency-coded SNNs to directly encode raw input images through temporal delays. First, the finite discrete formulation of the time dimension restricts the encoding resolution, which necessitates quantization of pixel intensity values and incurs substantial information loss. Second, latency coding scheme is particularly suited for representing sparse features due to its inherent alignment with sparse signal characteristics. It encodes feature saliency exclusively through precise spike timing—where higher intensity triggers an earlier spike—rather than relying on multiple spikes. Such efficiency not only preserves critical information but also mimics biological sensory systems that prioritize sparse, salient inputs through early-spike responses.\cite{2021A,2021Neural}

Consequently, we adopt a feature extraction layer to compute primary features from the input image. To facilitate intensity-to-latency encoding, we then rescale these feature values to the range [0, 1] using a sigmoid function. This preprocessing operation can be formulated as:
\begin{equation}
\boldsymbol{F}={\rm Sigmoid}(\Phi(\boldsymbol{I}))\in (0,1)^{C \times H \times W}
\end{equation}
where $\boldsymbol{F}$ denotes the primary features, $\boldsymbol{I}$ is the input image, $\Phi(\cdot)$ denotes the feature extraction layer, which includes the convolution operation and batch normalization \cite{2015Batch}.

Subsequently, the latency encoder $\text{LE}(\cdot)$ transforms the rescaled features $\boldsymbol F$ into spike trains $\boldsymbol X$ incorporating a temporal dimension, in which values determine the spike timing:
\begin{align}
\boldsymbol{X}=\text{LE}(\boldsymbol{F})&=[X[1],\cdots,X[T]] \in \{0,1\}^{T \times C \times H \times W}, \\
X_{c,h,w}[t]&=\left\{ \begin{aligned} &1,\;t=t_{s}(X_{c,h,w}[t]) \\
&0,\;otherwise
\end{aligned},\; \forall c,h,w \right. ,
\end{align}
where $X_{c,h,w}$ denotes the spike train of a single neuron in the feature map, $T$ is the maximum time step. Each neuron fires only once during this process, and $t_s(\cdot)$ defines the precise functional relationship between a neuron's activation value and its corresponding spike timing:
\begin{equation}
t_s(x) = \lceil(1 - x)T\rceil.
\end{equation}

Nevertheless, the operational principle of the latency encoder, which maps real-valued features to temporal spike sequences, inherently introduces a non-differentiable transformation and obstructs the gradient flow to earlier layers. To address this issue, we adopt the straight-through estimator (STE) to generate a surrogate gradient. The STE works by treating the non-differentiable function as an identity function during the backward pass of training. It allows the incoming gradient to be passed directly to the preceding layer, ignoring the differentiability of the function itself. Specifically, STE bypasses the problem of derivative calculation for the latency encoder by simply setting the gradient with respect to its input equal to the gradient with respect to its output. This process can be mathematically described as:
\begin{equation}
\frac{\partial L}{\partial \boldsymbol{F}}=\sum_t \frac{\partial L}{\partial \boldsymbol{X}[t]}.
\end{equation}

\subsection{Decoding with Membrane Potentials}

For the output layer of a latency-coded SNN, the classification is conventionally assigned to the category represented by the first-to-spike neuron. The timestamp of the first spike thereby dictates the network's overall inference latency. A practical complication, however, emerges in discrete-time implementations where temporal quantization can lead to the synchronous firing of multiple output neurons. This phenomenon obscures the identity of the true first-to-spike neuron, creating decisional ambiguity and potentially degrading classification accuracy.

To overcome this limitation and utilize the rich sub-threshold dynamics of the output neurons, we propose a refined decision-making policy that incorporates membrane potential as a tie-breaking principle. Let $\mathcal{T}$ denote the earliest spike time:
\begin{equation}
\mathcal{T}=\arg\min_t \sum_k S_k[t] > 0, \; t\in \{1,2,\cdots,T\},
\end{equation}

where at least one spike is emitted in the output layer. Then the policy operates as follows: in the case of a unique spike at $\mathcal{T}$, the prediction corresponds to the class of that sole neuron; in cases where a tie occurs — i.e., multiple neurons fire simultaneously at $\mathcal{T}$ — the ambiguity is resolved by comparing their membrane potentials before the spike. The neuron possessing the maximum membrane potential at time step $\mathcal{T}$ is considered the definitive winner, as it would be the first to reach the threshold in a continuous-time domain.

In summary, the final predicted class $\tilde{y}$ is the one corresponding to the neuron that exhibits the highest membrane potential at the earliest spike time $\mathcal{T}$, given by:
\begin{equation}
\tilde{y}=\arg\max_k U_k(\mathcal{T}),
\end{equation}
where $U_k(\mathcal{T})$ represents the membrane potential of the $k$-th output neuron at time $\mathcal{T}$. This method ensures a deterministic and unique decision by fully leveraging the analog state information inherent in the output neurons.

\subsection{Temporal Adaptive Decision}
To achieve both rapid and accurate decision-making, the output layer is trained to make a correct classification based on the first output spike. This is accomplished by balancing two learning objectives. On the one hand, for simple samples, the target neuron is encouraged to fire as early as possible to minimize inference latency. On the other hand, for more challenging samples, the network must be allowed sufficient time to integrate information across multiple timesteps to ensure high accuracy. Therefore, incorrect decisions made too early are penalized, forcing the network to perform inference over a longer time. To pursue the two objectives, we draw inspiration from concepts of conditional computation \cite{graves2016adaptive,banino2021pondernet} and propose temporal adaptive decision (TAD) loss function.

As discussed above, we expect the network to adapt its firing time to sample difficulty: easy samples should trigger an early spike, while hard samples should delay the decision to accumulate more evidence. To this end, at each timestep we compute the inverse entropy of the output logits as a confidence score, denoted as $\lambda[t]$: 
\begin{equation}
    \lambda[t]=1-\frac{H(\boldsymbol{z}[t])}{H_{max}}=1-\frac{-\sum_{i=1}^C z_i[t] \cdot log z_i[t]}{logC},
\end{equation}
where $H(\cdot)$ represents Shannon entropy, $\boldsymbol{z}[t]$ are normalized logits retrieved from the presynaptic current of the output layer $\boldsymbol{O}[t]$. This metric reflects the model's instantaneous uncertainty. When $\lambda[t]$ is high, the model is confident and thus more likely to fire early; when $\lambda[t]$ is low, the model remains uncertain and should continue integrating information, resulting in a lower firing tendency.

To further modulate the intensity of supervision signals across different timesteps, we apply a Softmax function with a temperature parameter $\mu$ to normalize $\lambda[t]$:
\begin{equation}
    \Lambda[t]=\frac{exp(\lambda[t]/\mu)}{\sum_{j=1}^T exp(\lambda[t]/\mu)}.
\end{equation}
During training, we weight the cross-entropy loss at each timestep by $\Lambda[t]$ and sum the weighted losses over time, yielding TAD loss: 
\begin{equation}
    \mathcal{L}_{TAD}=\sum_{t=1}^T\Lambda[t]\cdot \mathcal{L}_{CE}(\boldsymbol{O}[t],\boldsymbol{y}).
\end{equation}
Here, the temperature parameter $\mu$ controls the sharpness of the weighting distribution. A lower $\mu$ encourages the model to concentrate its supervision on the most confident timesteps (peak-seeking), while a higher $\mu$ results in a more uniform distribution of loss weights across time.

In practice, easy samples reach high confidence within the first few timesteps, whereas hard samples start with low confidence and increase it gradually over time. Consequently, easy samples receive stronger supervision at earlier timesteps, encouraging early spikes. And hard samples shift effective supervision signal to later timesteps, allowing longer inference time. This adaptive decision mechanism balances the trade-off between rapid and accurate decision-making.

\subsection{Gradient Flow Analysis}

In this section, we analyze the gradient flow advantage of latency coding, which provides richer gradient information than strict TTFS coding during BPTT. As described by the discrete LIF dynamics in Eqs. \eqref{eq:lif_membrane}--\eqref{eq:lif_reset}, each layer evolves through membrane integration, spike generation, and soft reset over time.

Since the Heaviside function $\Theta(\cdot)$ is non-differentiable, surrogate-gradient learning replaces its derivative with a surrogate function. We denote the surrogate derivative matrix by
\begin{equation}
\boldsymbol{\Psi}^l[t]
=
\frac{\partial \boldsymbol{S}^l[t]}{\partial \boldsymbol{U}^l[t]}.
\end{equation}
Because $\Theta(\cdot)$ is applied element-wise to each neuron, $\boldsymbol{\Psi}^l[t]$ is a diagonal matrix whose $i$-th diagonal entry is $\Psi_i^l[t]={\partial S_i^l[t]}/{\partial U_i^l[t]}$. The entries of $\boldsymbol{\Psi}^l[t]$ are nonzero mainly when the membrane potentials are close to the threshold $V_{th}$.

Let
\begin{equation}
\boldsymbol{g}^l[t]
=
\frac{\partial \mathcal{L}}{\partial \boldsymbol{U}^l[t]}
\end{equation}
be the gradient of the loss $\mathcal{L}$ with respect to the pre-reset membrane potential vector $\boldsymbol{U}^l[t]$. By BPTT, $\boldsymbol{g}^l[t]$ contains the local spike-induced gradient at timestep $t$ and the temporal gradient propagated from timestep $t+1$. The first term arises because $\boldsymbol{S}^l[t]$ depends on $\boldsymbol{U}^l[t]$ through the Heaviside function, and the second term arises because $\boldsymbol{U}^l[t+1]$ depends on $\boldsymbol{U}^l[t]$ through the membrane potential dynamics:
\begin{equation}
\boldsymbol{g}^l[t]
=
\frac{\partial \mathcal{L}}{\partial \boldsymbol{S}^l[t]}
\boldsymbol{\Psi}^l[t]
+
\boldsymbol{g}^l[t+1]
\frac{\partial \boldsymbol{U}^l[t+1]}{\partial \boldsymbol{U}^l[t]},
\end{equation}

Define the local spike-induced gradient source $\boldsymbol{e}^l[t]$ and the temporal Jacobian matrix $\boldsymbol{J}^l[t]$ as
\begin{equation}
\boldsymbol{e}^l[t]
\triangleq
\boldsymbol{\Psi}^l[t]
\frac{\partial \mathcal{L}}{\partial \boldsymbol{S}^l[t]},
\qquad
\boldsymbol{J}^l[t]
\triangleq
\frac{\partial \boldsymbol{U}^l[t+1]}{\partial \boldsymbol{U}^l[t]}.
\end{equation}
Then the gradient recursion can be simplified as
\begin{equation}
\boldsymbol{g}^l[t]
=
\boldsymbol{e}^l[t]
+
\boldsymbol{J}^l[t]\boldsymbol{g}^l[t+1].
\end{equation}
Expanding the recursion over time gives
\begin{equation}
\boldsymbol{g}^l[t]
=
\sum_{r=t}^{T}
\left(
\prod_{q=t}^{r-1}
\boldsymbol{J}^l[q]
\right)
\boldsymbol{e}^l[r],
\label{eq:vector_grad_unroll}
\end{equation}
where an empty product is defined as the identity matrix for the case $q>r-1$. Equation \eqref{eq:vector_grad_unroll} shows that the gradient at timestep $t$ receives contributions from all future spike-induced gradient sources.

We quantify the recovered gradient information from a neuron-wise perspective. Consider neuron $i$ in layer $l$, and let $\kappa_i^l$ denote its first-spike time. If neuron $i$ does not fire within the simulation window, we set $\kappa_i^l=\infty$. In strict TTFS coding, each neuron fires at most once, so all spike-induced gradient sources after $\kappa_i^l$ are zeroed out:
\begin{equation}
\left.e_i^l[t]\right|_{\mathrm{TTFS}}
=
\begin{cases}
e_i^l[t], & t \leq \kappa_i^l, \\[3pt]
0, & t > \kappa_i^l.
\end{cases}
\end{equation}
Latency coding restores these suppressed gradient sources by allowing intermediate neurons to fire multiple times, and every emitted spike contributes a valid surrogate-gradient term.
Taking the $i$-th component of Eq. \eqref{eq:vector_grad_unroll}, the additional gradient recovered by latency coding is
\begin{align}
\Delta g_i^l[t]
&=
g_i^l[t]
-
\left.g_i^l[t]\right|_{\mathrm{TTFS}}
\nonumber\\
&=
\sum_{r=\max(t,\kappa_i^l+1)}^{T}
\left(
\prod_{q=t}^{r-1}
J_i^l[q]
\right)
e_i^l[r],
\label{eq:neuron_lost_gradient}
\end{align}
where
$
J_i^l[q]
=
\partial U_i^l[q+1]/\partial U_i^l[q].
$
Here, we use the fact that for a standard feedforward LIF layer, the temporal dynamics are neuron-wise decoupled within the same layer, so the temporal Jacobian $\boldsymbol{J}^l[q]$ is diagonal with respect to neuron indices.

Equation \eqref{eq:neuron_lost_gradient} quantifies the extra gradient that latency coding provides over strict TTFS: it consists of all spike-induced gradient sources after the first-spike time $\kappa_i^l$, propagated back through the temporal Jacobians. Because $\kappa_i^l$ is neuron-specific, the benefit is heterogeneous across neurons—those that fire earlier recover more gradient terms, while those that fire later recover fewer.

For the synaptic weight $W_{ji}^l$ from presynaptic neuron $j$ to postsynaptic neuron $i$, the gradient is
\begin{equation}
\frac{\partial \mathcal{L}}{\partial W_{ji}^l}
=
\sum_{t=1}^{T}
\frac{\partial \mathcal{L}}{\partial U_{i}^l[t]}
\frac{\partial U_{i}^l[t]}{\partial W_{ji}^l}
=
\sum_{t=1}^{T}
g_i^l[t] S_j^{l-1}[t].
\label{eq:weight_gradient_neuron}
\end{equation}
This expression contains two factors: the postsynaptic backpropagated gradient $g_i^l[t]$ and the presynaptic spike $S_j^{l-1}[t]$.

From Eq. \eqref{eq:neuron_lost_gradient}, latency coding enriches the postsynaptic gradient with the additional term $\Delta g_i^l[t]$.
Thus, latency coding strengthens the postsynaptic gradient $g_i^l[t]$ with additional temporal contributions. Moreover, because intermediate neurons can fire multiple times, the presynaptic spike activity is also increased:
\begin{equation}
\left.\sum_{t=1}^{T}
S_j^{l-1}[t]\right|_{\mathrm{latency}}
\geq
\left.\sum_{t=1}^{T}
S_j^{l-1}[t]\right|_{\mathrm{TTFS}}.
\end{equation}
Thus, both terms in Eq. \eqref{eq:weight_gradient_neuron}, namely the backpropagated gradient $g_i^l[t]$ and the presynaptic spike $S_j^{l-1}[t]$, are enriched by latency coding.

Consequently, latency coding provides richer temporal gradient signals for weight updates than strict TTFS. By allowing multiple spikes in intermediate layers, it recovers gradient sources that TTFS discards and increases presynaptic spike activity, both of which help sustain effective gradient flow in deep SNNs.

\section{Experiments}

\subsection{Comparison to Related Works}

\begin{table*}[!t]
    \caption{Comparison of accuracy, inference time and sparsity on CIFAR-10, CIFAR-100, Tiny-ImageNet, ImageNet \cite{deng2009imagenet,ILSVRC15}, and CIFAR10-DVS datasets}
    \label{table_main}
    \centering
    \begin{threeparttable}
    \begin{tabular}{@{}ccccccc@{}}
    \toprule
    Dataset                     & Model                 & Architecture  & Method     & Accuracy(\%)   & Inference Time     & Sparsity      \\ \midrule
    \multirow{8}{*}{CIFAR-10}   & T2FSNN \cite{park2020t2fsnn}      & VGG-16        & Conversion & 91.43          & 680            & 0.25          \\
                                & TSC-SNN \cite{han2020deep}       & VGG-16        & Conversion & 93.63          & 2048           & 0.61          \\
                                & Park \textit{et al.} 2021 \cite{park2021training}      & VGG-16        & Direct     & 91.9           & 544            & 0.17          \\
                                & DTA-TTFS \cite{wei2023temporal}       & VGG-16        & Direct     & 93.05          & 160            & 0.26          \\
                                & Stanojevic \textit{et al.} 2024 \cite{stanojevic2024high} & VGG-16        & Conversion & 93.69          & 4096 per layer & 0.38          \\
                                & Ours                   & VGG-11         & Direct     & \textbf{93.60} & \textbf{1.00}  & \textbf{0.16} \\
                                & Ours                   & VGG-16         & Direct     & \textbf{93.12} & \textbf{1.13}  & \textbf{0.23} \\
                                & Ours                   & SEW-ResNet-18 & Direct     & \textbf{93.64} & \textbf{1.13}  & \textbf{0.22} \\ \midrule
    \multirow{7}{*}{CIFAR-100}  & T2FSNN \cite{park2020t2fsnn}      & VGG-16        & Conversion & 68.79          & 680            & 0.30          \\
                                & TSC-SNN \cite{han2020deep}       & VGG-16        & Conversion & 70.97          & 2048           & 0.61          \\
                                & Park \textit{et al.} 2021 \cite{park2021training}      & VGG-16        & Direct     & 65.98          & 544            & 0.28          \\
                                & DTA-TTFS \cite{wei2023temporal}       & VGG-16        & Direct     & 69.66          & 160            & 0.28          \\
                                & Stanojevic \textit{et al.} 2024 \cite{stanojevic2024high} & VGG-16        & Conversion & 72.24          & 4096 per layer & 0.38          \\
                                & Ours                   & VGG-11         & Direct     & \textbf{74.97} & \textbf{2.14}  & \textbf{0.52} \\
                                & Ours                   & SEW-ResNet-18 & Direct     & \textbf{74.80} & \textbf{3.92}  & \textbf{0.73} \\    \midrule
    \multirow{1}{*}{Tiny-ImageNet}& Ours                 & SEW-ResNet-18 & Direct    & \textbf{57.21}  & \textbf{4}  & \textbf{0.79}  \\  \midrule
    \multirow{1}{*}{ImageNet}& Ours                 & SEW-ResNet-34 & Direct    & \textbf{64.11}  & \textbf{4.00}  & \textbf{0.50}  \\  \midrule
    \multirow{2}{*}{CIFAR10-DVS}& T-SpikeFormer \cite{sun2025temporal} & T-Spikformer-2-256         & Direct    & 78.20  & 160  & -  \\
                                & Ours                   & VGG-11         & Direct    & \textbf{79.54}  & \textbf{4}  & \textbf{0.42}  \\  \bottomrule
    
    \end{tabular}
    \begin{tablenotes}
        \item $^a$ Method: Conversion denotes ANN-to-SNN conversion methods, while Direct denotes directly trained methods.
        \item $^b$ Inference Time: the average number of discrete simulation timesteps used for inference.
        \item $^c$ Sparsity: calculated as the total number of spikes divided by the product of the total number of neurons and timesteps.
    \end{tablenotes}
    \end{threeparttable}
\end{table*}

\subsubsection{Experimental Setup}
In this section, we benchmark our proposed latency coding method on several image classification datasets, covering both static images and neuromorphic data, and compare our results with previous TTFS coding methods.

We employ four network architectures: VGG-11, VGG-16 \cite{simonyan2014very}, SEW-ResNet-18 \cite{fang2021deep}, and SEW-ResNet-34. For training efficiency, we remove the two fully connected layers in VGG-11 and VGG-16. For all datasets except ImageNet, we use the AdamW optimizer \cite{loshchilov2017decoupled} with an initial learning rate of 0.001 and a cosine decay schedule reducing to 0, training for 200 epochs on a single NVIDIA RTX 4090 GPU. For ImageNet, we use the SGD optimizer with an initial learning rate of 0.1, and train for 120 epochs on a single NVIDIA A100 GPU. In all experiments, $\mu$ is set to 2. The comprehensive comparison results are summarized in Table \ref{table_main}.

\subsubsection{CIFAR-10 \& CIFAR-100}
We first apply our latency coding framework to the CIFAR-10 and CIFAR-100 datasets, adopting AutoAugment \cite{cubuk2019autoaugment} as the data augmentation technique.

As shown in Table \ref{table_main}, our method achieves a significant breakthrough in inference latency while maintaining competitive or superior accuracy compared to existing temporal coding SNNs. On CIFAR-10, our VGG-11 model achieves 93.60\% accuracy with an inference time of only 1.00 timesteps and a sparsity of 0.16. This is a remarkable improvement in efficiency compared to \cite{stanojevic2024high}, which requires up to 4096 timesteps per layer to achieve a comparable accuracy of 93.69\%. Furthermore, compared to DTA-TTFS \cite{wei2023temporal} (93.05\% accuracy, 160 timesteps), our method not only improves accuracy by 0.55\% but also reduces inference latency by two orders of magnitude.

On the more challenging CIFAR-100 dataset, our method demonstrates clear superiority. Our VGG-11 and SEW-ResNet-18 models achieve 74.97\% and 74.80\% accuracy, respectively, significantly outperforming the best baseline (72.24\%). Notably, this performance gain is achieved with ultra-low latency (2 to 4 timesteps), whereas other methods typically require hundreds or thousands of timesteps (680 timesteps for T2FSNN and 2048 timesteps for TSC-SNN). The results validate that our proposed TAD training strategy successfully mitigates the latency bottleneck inherent in temporal coding SNNs without compromising performance.

\subsubsection{Tiny-ImageNet \& ImageNet}
To verify the scalability of our method on larger-scale datasets, we evaluate it on Tiny-ImageNet \cite{le2015tiny}, which consists of 200 classes with 64$\times$64 resolution images, and ImageNet \cite{deng2009imagenet,ILSVRC15}, which provides a more challenging large-scale benchmark with 1000 object categories. We employ RandAugment \cite{cubuk2020randaugment} for data augmentation.

As reported in Table \ref{table_main}, our SEW-ResNet-18 model achieves an accuracy of 57.21\% on Tiny-ImageNet with an inference time of 4 timesteps and a sparsity of 0.79. On ImageNet, the SEW-ResNet-34 model obtains 64.11\% accuracy with an inference time of 4.00 timesteps and a sparsity of 0.51. The results further demonstrates that the proposed latency coding framework can be extended to large-scale visual recognition while preserving the ultra-low-latency inference property.

\subsubsection{CIFAR10-DVS}
We further evaluate our method on CIFAR10-DVS \cite{li2017cifar10}, which is considered one of the most challenging mainstream neuromorphic datasets. CIFAR10-DVS converts static images into event streams, capturing temporal dynamics. However, since it provides only 900 training samples per category, large models are prone to overfitting. Therefore, we adopt a lightweight VGG-11 architecture for this task and use \cite{li2022neuromorphic} as the augmentation method.

As shown in Table \ref{table_main}, we achieve 79.54\% accuracy, outperforming the recent Transformer-based SNN, T-SpikeFormer \cite{sun2025temporal} (78.20\%), by 1.34\%. Crucially, our model operates with an inference time of only 4 timesteps, which is $40\times$ faster than the 160 timesteps required by T-SpikeFormer. This result confirms that our latency coding framework effectively leverages the temporal information in event-based data while maximizing inference speed.

\subsection{Energy Estimation}
To provide a rigorous and comprehensive evaluation on the energy efficiency of our model, we quantify energy consumption using two distinct methodologies.

Following common practices in neuromorphic computing literature \cite{rathi2021diet,yao2024spike}, we first estimate the theoretical energy consumption based on operation counts. 
For ANNs, computational costs are usually calculated by the number of floating-point operations (FLOPs), which consist primarily of MAC operations \cite{molchanov2016pruning}. Similarly, for SNNs, energy consumption can also be evaluated by another operation measurement, synaptic operations (SOPs), which were initially proposed by IBM TrueNorth team \cite{merolla2014million} and adopted by numerous studies to benchmark hardware efficiency \cite{bu2023optimal, chen2023hybrid}. Despite the fact that estimation based on operation counts neglects the hardware architecture and the static power consumption, it is still a simple but practical tool to evaluate algorithms' performance. We select this method to analyze the energy consumption of our model and compare it to ANN and rate-based SNN.

Specifically, the energy cost of ANN ($E_{ANN}$) and SNN ($E_{SNN}$) in a single inference can be calculated as:
\begin{equation}
    E_{ANN}=\sum_l N_{FLOPs}^l \cdot E_{MAC},
\end{equation}
\begin{equation}
    E_{SNN} = E_{MAC} \cdot N_{FLOPs}^1 + E_{AC} \cdot \sum_{l=2}^L N_{SOPs}^l.
\end{equation}
For $E_{ANN}$, $N_{FLOPs}^l$ denotes the times of FLOPs in $l$-th layer and $E_{MAC}$ is the energy cost of MAC operation. 
For $E_{SNN}$, calculation is a little different, where $N_{SOPs}^l$ denotes the number of SOPs  in $l$-th layer and $E_{AC}$ is the energy cost of AC operation. Since the first layer adopts direct coding to process continuous-valued inputs, its computational cost is quantified the same as ANN, while subsequent layers are evaluated using SOPs. In practice, the number of SOPs is computed as $N_{SOPs}^l = \alpha^{l-1} \cdot N_{FLOPs}^l$. Here, $\alpha^{l-1}$ denotes the firing rate of the previous layer, which equals the ratio of activated synapses between $(l-1)$-th and $l$-th layer.
For a convolutional layer, the times of FLOPs is determined by the output feature map dimensions and the kernel configuration. Let $H_{out}$ and $W_{out}$ denote the height and width of output feature maps, $C_{in}$ and $C_{out}$ the number of input and output channels, $K$ the kernel size. Then total FLOPs are calculated as $N_{FLOPs\_conv} = H_{out} \cdot W_{out} \cdot C_{in} \cdot C_{out} \cdot K^2$. 
In a fully connected layer, every input neuron is connected to every output neuron. If $I$ represents the dimension of input and $O$ represents the dimension of output, the total FLOPs is given by $N_{FLOPs\_fc} = I \cdot O$.
According to previous works \cite{yao2024spike, yin2021accurate, panda2020toward}, we adopt the energy data from 45nm CMOS technology , where a 32-bit floating-point accumulation ($E_{AC}$) consumes 0.9 pJ and a 32-bit floating-point multiply-accumulate ($E_{MAC}$) consumes 4.6 pJ \cite{horowitz20141}.

We benchmark the energy consumption of our model on CIFAR-10 and CIFAR-100 dataset. All SNNs use SEW-ResNet-18 architecture and are trained with 4 timesteps. As shown in Table \ref{table_energy}, the proposed latency-coded SNN achieves state-of-the-art energy efficiency with comparable accuracy to ANN and rate-coded SNN. Specifically, on the CIFAR-10 dataset, our model consumes only 0.097 mJ, corresponding to a mere 4.1\% of the energy required by the ANN counterpart. Furthermore, compared to the rate-coded SNN, our method achieves more than 3$\times$ reduction in energy consumption. On the more complex CIFAR-100 dataset, our model maintains its efficiency advantage, consuming only 12.6\% of the ANN energy. Crucially, while surpassing the performance of rate-coded SNN, our approach simultaneously realizes an additional 2.5\% reduction in energy consumption. Although a minor accuracy drop was observed on CIFAR-10, the proposed model provides a superior energy efficiency, making it highly suitable for resource-constrained edge computing scenarios.

\begin{table}[!t]
    \caption{Comparison of Theoretical Energy Consumption\label{table_energy}}
    \centering
    \begin{threeparttable}
    \begin{tabular}{@{}ccccc@{}}
    \toprule
    Dataset                    & Model      & Energy(mJ)    & Ratio  &$\Delta$Acc.(\%)       \\ \midrule
    \multirow{3}{*}{CIFAR-10}  & ANN$^*$        & 2.395         & 100\%   & -               \\
                               & Rate\cite{deng2022temporal}      & 0.339         & 14.2\%   & -0.37          \\
                               & \textbf{Latency}    & \textbf{0.097}         & \textbf{4.1\%}    & -1.37                \\ \midrule                         
    \multirow{3}{*}{CIFAR-100} & ANN$^*$      & 2.395         & 100\%       & -           \\
                               & Rate\cite{deng2022temporal}          & 0.362         & 15.1\%    & -0.80             \\
                               & \textbf{Latency}         & \textbf{0.302}         & \textbf{12.6\%}  & -0.55           \\ \bottomrule
    \end{tabular}
    \begin{tablenotes}
        \item $^*$ Self-implemented results. Ratio = SNN Energy / ANN Energy. $\Delta$Acc. = SNN Accuracy - ANN Accuracy
    \end{tablenotes}
    \end{threeparttable}
\end{table}

While methods based on operation counts reflect computational costs, they overlook the static energy consumption that is highly related to simulation duration. To provide a more holistic perspective, particularly highlighting the advantages of our low-latency design, we employ analysis from \cite{park2020t2fsnn} that accounts for inference latency, spike counts and characteristics of neuromorphic architectures. Specifically, this method uses the formula:
\begin{equation}
    E_{SNN} = E_{Static} \times Timesteps + E_{Dynamic} \times Spikes, 
\end{equation}
where $E_{Static}$ and $E_{Dynamic}$ represent static and dynamic energy coefficients, respectively, which are determined by neuromorphic architectures. We conduct evaluation based on TrueNorth \cite{merolla2014million} and SpiNNaker \cite{furber2014spinnaker} architectures. [$E_{Static}$, $E_{Dynamic}$] are set to [0.6, 0.4] and [0.36, 0.64] for TrueNorth and SpiNNaker, respectively, according to \cite{park2020t2fsnn}.

With this approach, we compare our model with other TTFS methods on CIFAR-10 using VGG-16 architecture. Table \ref{table_energy_2} presents the normalized energy (values of $Timesteps$ and $Spikes$ are normalized before calculating results) estimated for TrueNorth and SpiNNaker platforms.
On the TrueNorth platform, which has a relatively high static power proportion, our model achieves a normalized energy of 0.311, roughly 30\% of the baseline and significantly lower than the high-latency method (1.530). On SpiNNaker, our model records the lowest normalized energy of 0.494. Remarkably, our method requires only 4 timesteps, which is orders of magnitude lower than other TTFS approaches (ranging from 160 to 1024 timesteps). In addition, it generates the fewest spikes (5.3$\times$10$^4$), indicating high sparsity. By minimizing both terms of the formula, our model outperforms existing TTFS methods in terms of energy efficiency.

\begin{table}
    \caption{Normalized Energy \label{table_energy_2}}
    \centering
    \begin{tabular}{@{}cccccc@{}}
    \toprule
    Neural                       & Time           & Spikes             & \multicolumn{1}{c|}{Accuracy}                      & \multicolumn{2}{c}{Normalized Energy}       \\ 
    Coding                       & Steps          & ($10^4$)           & \multicolumn{1}{c|}{(\%)}           & TrueNorth      & SpiNNaker \\  \midrule
    TTFS\cite{park2020t2fsnn}    & 680            & 6.9              & \multicolumn{1}{c|}{91.43}          & 1$\times$        & 1$\times$      \\   
    TTFS\cite{park2021training}  & 544            & 6.7              & \multicolumn{1}{c|}{91.90}          & 0.868$\times$          & 0.909$\times$       \\
    TTFS\cite{wei2023temporal}   & 160            & 7.3              & \multicolumn{1}{c|}{93.05}          & 0.564$\times$          & 0.762$\times$       \\
    TTFS\cite{stanojevic2024high}$^*$ & 1024      & 10.8              & \multicolumn{1}{c|}{93.64}          & 1.530$\times$          & 1.544$\times$      \\
    \textbf{Latency}             & \textbf{1.31}     & \textbf{6.3}     & \multicolumn{1}{c|}{\textbf{93.12}} & \textbf{0.366$\times$} & \textbf{0.492$\times$}      \\  \bottomrule

    \end{tabular}
    \begin{tablenotes}
        \item $^*$ Choose timesteps with which the model has a comparable accuracy.
    \end{tablenotes}
\end{table}

\subsection{Robustness}

To evaluate the robustness of the proposed latency-coded SNN against sensory perturbations, we conducted a comparative analysis with a rate-coded SNN. For this assessment, we employed the widely used CIFAR-10-C and CIFAR-100-C benchmarks, originally proposed in \cite{hendrycks2019benchmarking}. These datasets are constructed by applying a comprehensive suite of algorithmically generated, realistic corruptions to the clean images of the original CIFAR test sets. Specifically, the benchmarks select 15 distinct corruption types—categorized into groups such as noise, blur, weather, and digital artifacts—with each type stratified across five increasing levels of severity. This rigorous setup allows for a quantification of model stability under varying degrees of image degradation. In this section, we employ the SEW-ResNet-18 architecture as the backbone for all experiments and select Temporal Efficient Training (TET) method as the representative rate-coded SNN baseline. Both models are trained with 4 timesteps.

Table \ref{table_robust} presents comprehensive results between the rate-coded and the proposed latency-coded SNNs across CIFAR-10-C and CIFAR-100-C datasets. First, Regarding classification performance on clean data, the proposed latency coding scheme maintains an accuracy comparable to the rate coding baseline. On CIFAR-100, two coding schemes achieve nearly identical performance (25.4\% vs. 25.2\%). On CIFAR-10, clean error of latency coding is slightly higher (5.4\% vs. 6.4\%). However, a significant divergence is observed in terms of robustness. The latency-coded SNN consistently outperforms the rate-coded counterpart across both benchmarks. On CIFAR-100-C, our approach yields a 50.8\% mCE, a substantial 6.1\% improvement over the rate-coded model (56.9\%), with lower error rates on most corruption types. Likewise, on CIFAR-10-C, the latency-coded model decreases the mCE by 0.6\% and outperforms the baseline on over half of the corruptions.

\begin{figure*}
    \centering
    \includegraphics[width=1\linewidth]{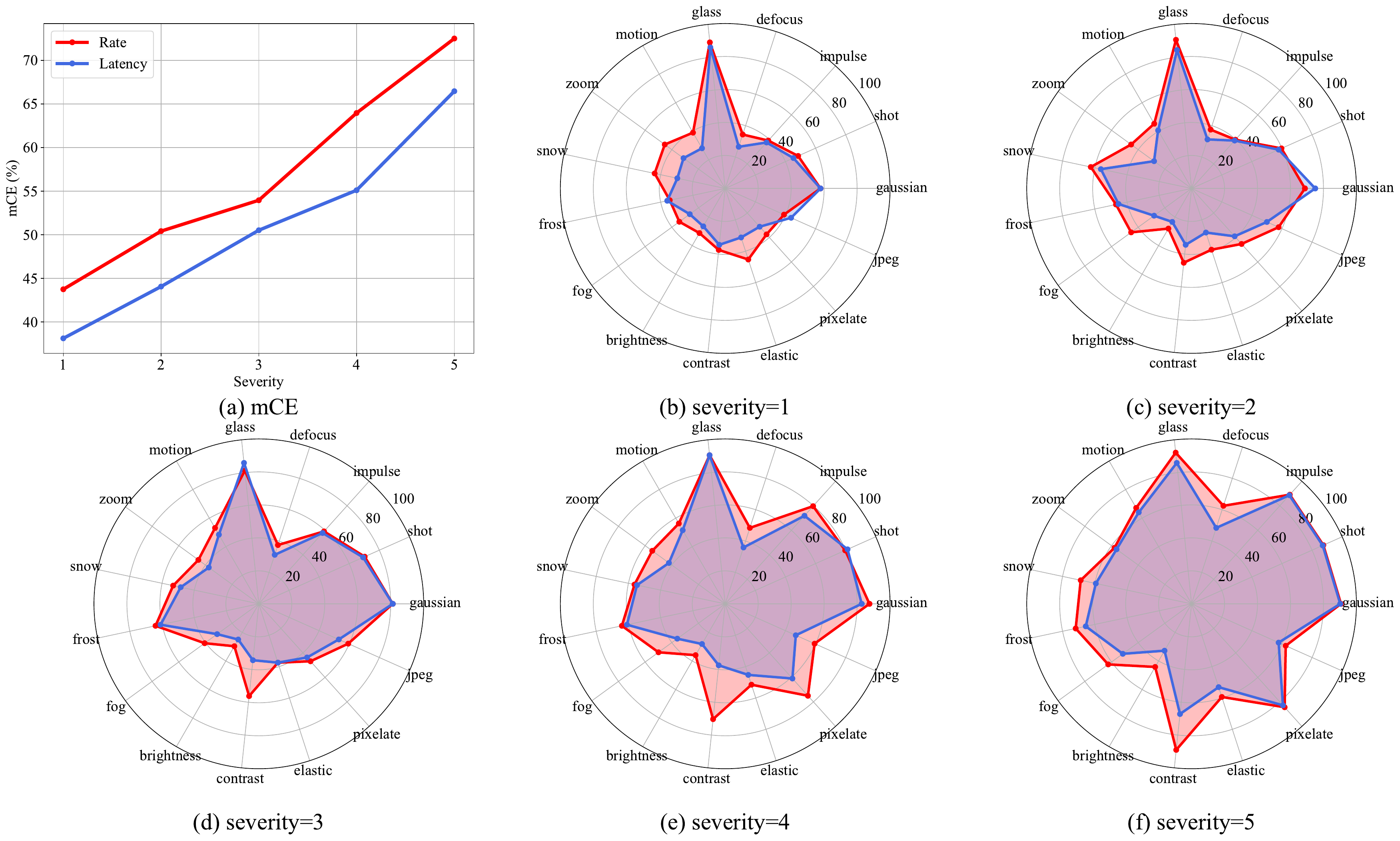}
    \caption{Performance evolution across severity levels on CIFAR-100-C. (a) Comparison of mCE from Severity 1 to 5. (b)-(f) Radar charts illustrating the error rates for 15 specific corruption types at each severity. Smaller polygon area indicates better robustness.}
    \label{robust}
\end{figure*}

Fig. \ref{robust} provides a more detailed visualization of models' performance on the CIFAR-100-C dataset across all severity levels. Ploted in Fig. \ref{robust}(a) is the mCE against five severity levels. As expected, the error rate for both coding schemes monotonically increase with the severity of corruption. Nevertheless, the latency-coded SNN (blue line) consistently maintains a lower error trajectory compared to the rate-coded baseline (red line) across all spectrum. Notably, even at the highest severity level (severity=5), where image features are heavily distorted, the latency model still retains a performance margin, demonstrating stronger inherent stability against input perturbations.

The radar charts in Fig. \ref{robust}(b)-(f) further illustrate the error distribution over different corruption types at each severity level. In these plots, a smaller polygon area indicates lower error rates and better robustness. As shown in the figures, the blue polygon (Latency) is generally enclosed within the red polygon (Rate), which signifies that the latency coding scheme performs better for most corruption types. 

At lower severity levels, the latency coding scheme shows clear advantages in resisting certain corruptions. It exhibits strong robustness against \textit{Defocus blur}, \textit{Brightness}, and \textit{Elastic} transformations, where the performance gap over the rate-coded baseline is particularly pronounced. 
As the corruption severity increases, our method exhibits a slower degradation, maintaining stability against \textit{Fog}, \textit{Snow}, \textit{Zoom blur}, \textit{JPEG} compression, and \textit{Contrast} changes. In these categories, the latency-coded model consistently outperforms the rate-coded baseline, even under extreme conditions. However, latency coding is less effective against \textit{Gaussian noise} and \textit{Shot noise}, where its error rates are higher than those of rate coding in several severity levels. This suggests that while the latency coding scheme is more resilient to structural distortions, weather-related artifacts, and blurs, it remains highly sensitive to certain pixel-level statistical noise distributions like Gaussian noise.

\begin{table*}[]
    \centering
\resizebox{\linewidth}{!}{
\begin{threeparttable}
\caption{Robustness comparison between Rate-coded and Latency-coded SNNs on CIFAR-10-C and CIFAR-100-C datasets \label{table_robust}}

\begin{tabular}{@{}cccccccccccccccccccc@{}}
\multicolumn{1}{l}{} & \multicolumn{1}{l}{} & \multicolumn{1}{l}{}  & \multicolumn{1}{l}{} & \multicolumn{3}{c}{Noise}                    & \multicolumn{4}{c}{Blur}                             & \multicolumn{4}{c}{Weather}                      & \multicolumn{4}{c}{Digital}       \\ \toprule
Dataset                     & Coding Scheme        & Clean                 & \multicolumn{1}{c|}{\textbf{mCE}}          & Gauss & Shot & \multicolumn{1}{c|}{Impulse} & Defocus & Glass & Motion & \multicolumn{1}{c|}{Zoom} & Snow & Frost & Fog & \multicolumn{1}{c|}{Bright} & Contrast & Elastic & Pixel & JPEG \\ \midrule
\multirow{2}{*}{CIFAR-10}   & Rate \cite{deng2022temporal}                 & 5.4                   & \multicolumn{1}{c|}{34.7}                  & 58.8     & \textbf{44.4}   & \multicolumn{1}{c|}{45.6}      & \textbf{21.9}      & 66.6    & 32.5     & \multicolumn{1}{c|}{31.2}   & \textbf{25.6}   & \textbf{28.4}    & 20.6  & \multicolumn{1}{c|}{\textbf{8.4}}       & 39.7         & 31.9        & 39.4    & \textbf{25.9}   \\
                            & Latency              & 6.4                   & \multicolumn{1}{c|}{\textbf{33.9}}         & \textbf{58.1}     & 45.0   & \multicolumn{1}{c|}{\textbf{41.9}}                              & 22.5      & \textbf{65.6}    & \textbf{30.3}                             & \multicolumn{1}{c|}{\textbf{29.1}}   & 29.4   & 31.2    & \textbf{17.8}  & \multicolumn{1}{c|}{12.2}     & \textbf{31.9}        & \textbf{27.8}       & \textbf{37.2}    & 28.8   \\ \midrule
\multirow{2}{*}{CIFAR-100}  & Rate \cite{deng2022temporal}                & 25.4                  & \multicolumn{1}{c|}{56.9}                  & \textbf{77.2}     & 69.1   & \multicolumn{1}{c|}{61.3}     & 44.1      & 88.8    & 52.2     & \multicolumn{1}{c|}{49.7}   & 56.9   & 56.2    & 46.6  & \multicolumn{1}{c|}{33.8}     & 59.7       & 46.6      & 57.8    & 55.6   \\
                            & Latency              & 25.2                  & \multicolumn{1}{c|}{\textbf{50.8}}         & 77.4     & \textbf{68.1}   & \multicolumn{1}{c|}{\textbf{59.0}}                            & \textbf{34.7}      & \textbf{86.6}        & \textbf{46.6}                             & \multicolumn{1}{c|}{\textbf{39.1}}   & \textbf{49.7}   & \textbf{53.8}    & \textbf{34.7}  & \multicolumn{1}{c|}{\textbf{27.2}}    & \textbf{41.6}      & \textbf{39.1}     & \textbf{51.6}    & \textbf{50.3}   \\ \bottomrule
\end{tabular} 

\begin{tablenotes}
    \item[] Values indicate classification error rate. The mCE is mean corruption error, the Clean is performance on clean data and the rest are error rates on each type of corruptions averaged on 5 serveritis. Bold values indicate lower corruption error and better robustness.
\end{tablenotes}
\end{threeparttable}
}
\end{table*}

\subsection{Temporal Similarity of Spike Representation}

\begin{figure}
    \centering
    \includegraphics[width=1\linewidth]{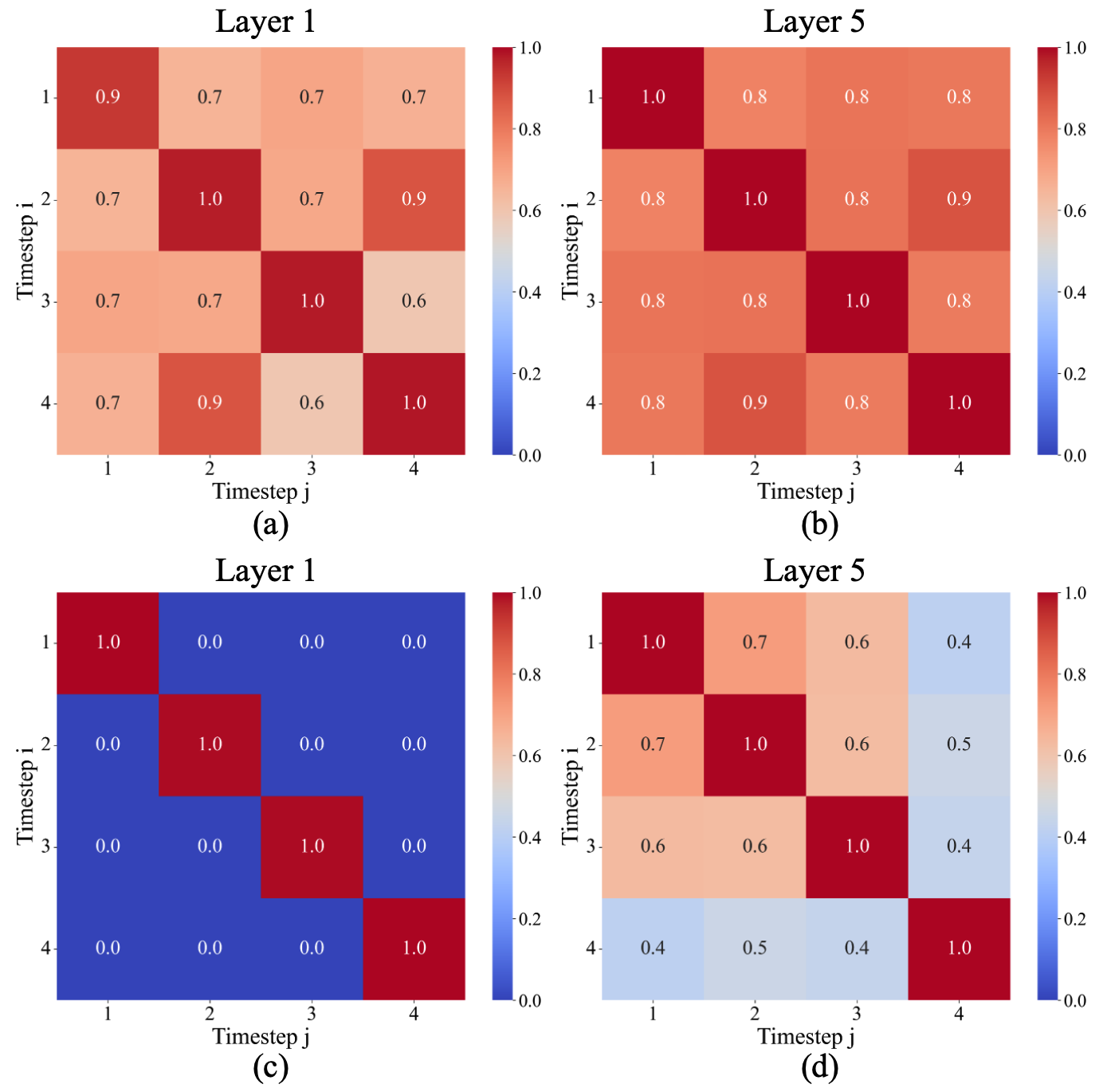}
    \caption{Temporal Similarity Matrix. Calculated on CIFAR100 test set. (a)(b) Rate-coded SNN. (c)(d) Latency-coded SNN.}
    \label{similarity}
\end{figure}

To investigate the temporal dynamics of different encoding schemes, we employed time-resolved representation similarity analysis on the CIFAR-100 dataset. To quantitatively evaluate the temporal correlation of neural activity, we define the temporal similarity matrix $M$. Let $\mathcal{S}_k[t] \in \{0,1\}^{C\times H \times W}$ denote the spike feature map of the $k$-th sample at timestep $t$, where $C,H,W$ are the number of channels, height, and width, respectively. To facilitate the calculation, each feature map is flattened into single feature vector $\boldsymbol{v}_k[t]=\text{flatten}(\mathcal{S}_k[t])\in \{0,1\}^D$, where $D=C \cdot H \cdot W$. Each element $M_{i,j}$ represents the average cosine similarity between the spike features at timestep $t_i$ and $t_j$ across the entire dataset of $N$ samples:
\begin{equation}
M_{i,j} = \frac{1}{N} \sum_{k=1}^N \frac{\boldsymbol{v}_k[t_i]\cdot \boldsymbol{v}_k[t_j]}{\|\boldsymbol{v}_k[t_i]\| \cdot \|\boldsymbol{v}_k[t_j]\|},
\end{equation}
where $\|\cdot \|$ denotes $L_2$ norm. This matrix captures the degree of overlap in spikes patterns across the temporal dimension. Here, we calculate and visualize the temporal similarity matrix for layer 1 and layer 5. These shallow layers function as SNNs' counterpart to the retina's preliminary signal processing, forming the foundation for all subsequent spike-based computations. Therefore, the quality and properties of the initial features are crucial to the network's performance.

As illustrated in Fig. \ref{similarity}(a) and (b), the rate-coded SNN exhibits high similarity values (0.7-1.0) across nearly all timestep pairs. These high off-diagonal values indicate that the activated feature patterns are persistent and highly redundant, suggesting a relatively static representation where the temporal dimension is underutilized. In contrast, the latency-coded SNN (Fig. \ref{similarity}(c) and (d)) reveals a much sparser similarity profile. In Layer 1, the matrix is strictly diagonal, meaning the set of activated neurons at any given time is entirely distinct from others. Even as features integrate in layer 5, the inter-step similarity remains significantly lower than that of the rate-coded counterpart.

This low temporal similarity offers two primary advantages. First, it maximizes temporal expressive ability by ensuring that different information is encoded at different moments, allowing for highly dynamic and non-overlapping neural activity. Second, the decorrelation in shallow layers filters out transient noise and prevent input perturbations from propagating through the temporal axis. This phenomenon may explain the better robustness of latency-coded SNNs compared to rate-coded SNNs.

\subsection{Validation Study}

\subsubsection{LE Module}

\begin{figure}
    \centering
    \includegraphics[width=0.9\linewidth]{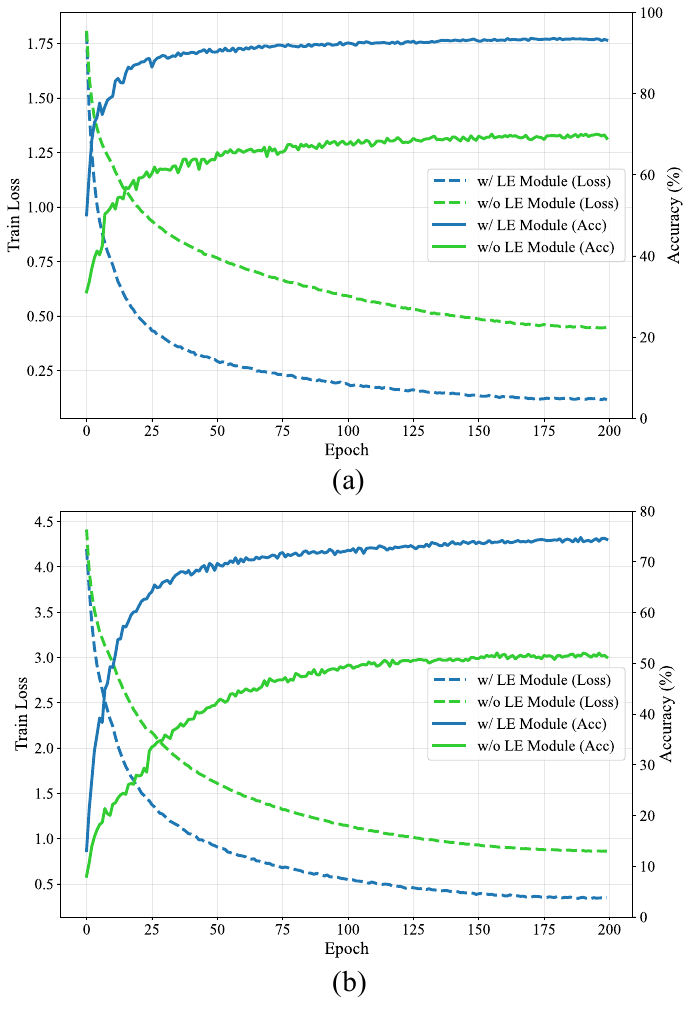}
    \caption{Ablation study of the proposed LE module on (a) CIFAR-10 and (b) CIFAR-100.}
    \label{compare}
\end{figure}

We further investigate the impact of the input encoding strategy by comparing the LE module with a version that encodes raw image intensities directly into the spike latency. As illustrated in \ref{compare}, a substantial performance gap exists between the two configurations across both datasets. The model without the LE module (green curves) exhibits significantly higher training loss and a much lower accuracy plateau, indicating a severe bottleneck in the learning process. This empirical evidence confirms that direct pixel-to-latency mapping results in information loss problem, where the network fails to capture sufficient discriminative features due to the limited resolution of the temporal dimension.

The performance degradation in the baseline stems from the mismatch between the dense intensity of raw images and the discrete nature of latency coding. Within a finite time dimension, direct encoding necessitates a coarse quantization of pixel values into a few discrete timesteps, which inevitably discards fine-grained visual details essential for high-level recognition. Furthermore, latency coding is inherently optimized for representing sparse, salient features rather than dense raw signals. Forcing raw pixels into a temporal-sparse format without prior feature refinement leads to a degraded input representation that impedes gradient flow and restricts the network's expressive capacity.

By transforming dense inputs into salient feature maps, the LE module aligns the input characteristics with the early-spike mechanism of latency coding, mimicking biological sensory systems that prioritize salient signals. These results demonstrate that the LE module is not merely an auxiliary component but a critical architectural requirement for latency-coded SNNs to achieve robust convergence and superior classification performance.

\subsubsection{TAD loss}

In order to validate the effectiveness of TAD loss, we perform ablation studies on the CIFAR-10 and CIFAR-100 datasets using the SEW-ResNet-18 architecture. Table \ref{table_loss} presents the ablation studies concerning the proposed Temporal Adaptive Decision (TAD) loss. For the baseline (w/o TAD) case, models are optimized using the vanilla Cross-Entropy loss: $\mathcal{L}=\text{CrossEntropy}(\frac{1}{T}\sum_{t=1}^T \boldsymbol{O}[t], \boldsymbol{y})$. The quantitative results confirm the effectiveness of our design philosophy: minimizing the temporal cost required to achieve high-precision classification. Instead of compromising accuracy for speed, the TAD loss guides the network to make correct decisions using the minimal necessary timesteps. As shown in Table \ref{table_loss}, the model trained with TAD loss consistently outperforms the baseline in accuracy across both datasets. Specifically, on CIFAR-10 dataset, it achieves higher accuracy (from 92.75\% to 93.64\%) with reduced latency (from 1.27 to 1.14 timesteps), proving that the proposed loss successfully eliminates redundant waiting times for confident predictions. For the more complex CIFAR-100 dataset, it gains 1.49\% (from 73.13\% to 74.80\%). The slight increase in timesteps reflects the network's adaptive behavior—investing necessary additional time to ensure correctness for hard samples, rather than making hasty, incorrect predictions. 

\begin{table}[!t]
    \caption{Ablation studies of TAD loss\label{table_loss}}
    \centering
    \begin{threeparttable}
    \begin{tabular}{@{}ccccc@{}}
    \toprule
    Dataset                    & Method      & Acc.(\%)        & Timesteps    & Sparsity     \\ \midrule
    \multirow{2}{*}{CIFAR-10}  & w/o TAD     & 92.75               & 1.27    & 0.17           \\
                               & w/ TAD      & \textbf{93.64}      & \textbf{1.13}     & \textbf{0.16}           \\ \midrule                         
    \multirow{2}{*}{CIFAR-100} & w/o TAD     & 73.31               & 3.26   & 0.58        \\
                               & w/ TAD      & \textbf{74.80}      & \textbf{3.92}  & \textbf{0.73}     \\ \bottomrule
    \end{tabular}
    \begin{tablenotes}
        \item 
    \end{tablenotes}
    \end{threeparttable}
\end{table}

\begin{figure}
    \centering
    \includegraphics[width=0.9\linewidth]{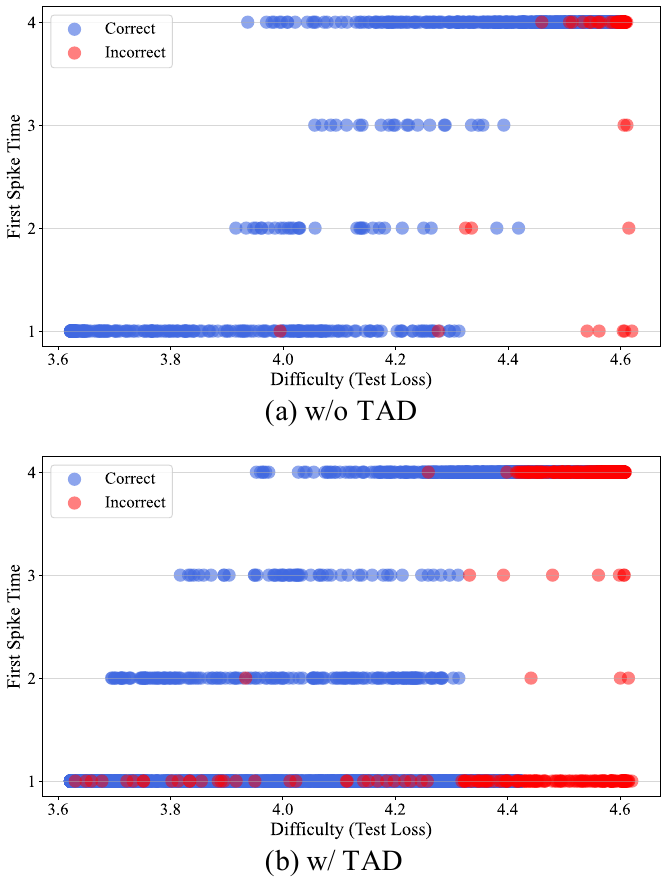}
    \caption{Distribution of first spike time against sample difficulty. (a) Training with TAD loss. (b) Training with vanilla loss.}
    \label{loss}
\end{figure}

To further verify whether TAD loss effectively guides the network to adjust the first spike time adaptively according to sample difficulty, we conducted a analysis of the per-sample inference time on the CIFAR-100 test set. Fig. \ref{loss} visualizes the distribution of the first spike time against sample difficulty. To ensure a fair quantificated comparisom, we utilize the same vanilla cross-entropy loss as the difficulty metric. In the scatter plots, blue points denote correctly classified samples, while red points represent incorrect samples. As illustrated in Fig. \ref{loss}(a), the model trained with TAD loss exhibits a pronounced positive correlation between the first spike time and sample difficulty.

Firstly, the distribution of misclassified samples (red points) highlights a critical advantage of TAD loss. In the baseline model (Fig. \ref{loss}(b)), a significant number of ``hard'' samples (high difficulty) appear in the bottom-right region. This indicates that the baseline model often makes hasty, incorrect predictions at an early timestep ($T=1$) without accumulating sufficient evidence. Conversely, the errors in the TAD-trained model are concentrated in the top-right corner ($T=4$). This shift suggests that for challenging samples, the model learns to inhibit premature firing, forcing itself to integrate the full temporal information. Although some of these samples remain misclassified, the network correctly identifies them as ``hard'' and attempts to maximize decision fidelity.

Secondly, regarding correctly classified samples (blue points), the TAD loss effectively eliminates computational redundancy. In Fig. \ref{loss}(a), the region corresponding to low-difficulty samples at intermediate timesteps is notably sparse compared to the baseline, which indicates that the TAD-trained model adaptively pushes the first spike time of easy samples to an early timestep of $T=1$. In contrast, the baseline model (Fig. \ref{loss}(b)) often processes easy samples using $T=2$ or $T=3$, incurring unnecessary latency. Admittedly, TAD loss will lead to a distribution somewhat clustered at the extremes ($T=1$ and $T=4$), suggesting further refinement of the reweighting strategy.

In conclusion, these observations explain the macroscopic results shown in Table \ref{table_loss}. Although the average latency on CIFAR-100 increases slightly, this is not due to inefficiency, but rather an investment of time into harder samples to boost accuracy. The TAD loss successfully reshapes the temporal inference trajectory and aligns with the ideal inference strategy: minimizing latency for clear signals while reserving computational resources for ambiguous inputs.

\subsubsection{Time Scalability Robustness}

\begin{figure*}
    \centering
    \includegraphics[width=1\linewidth]{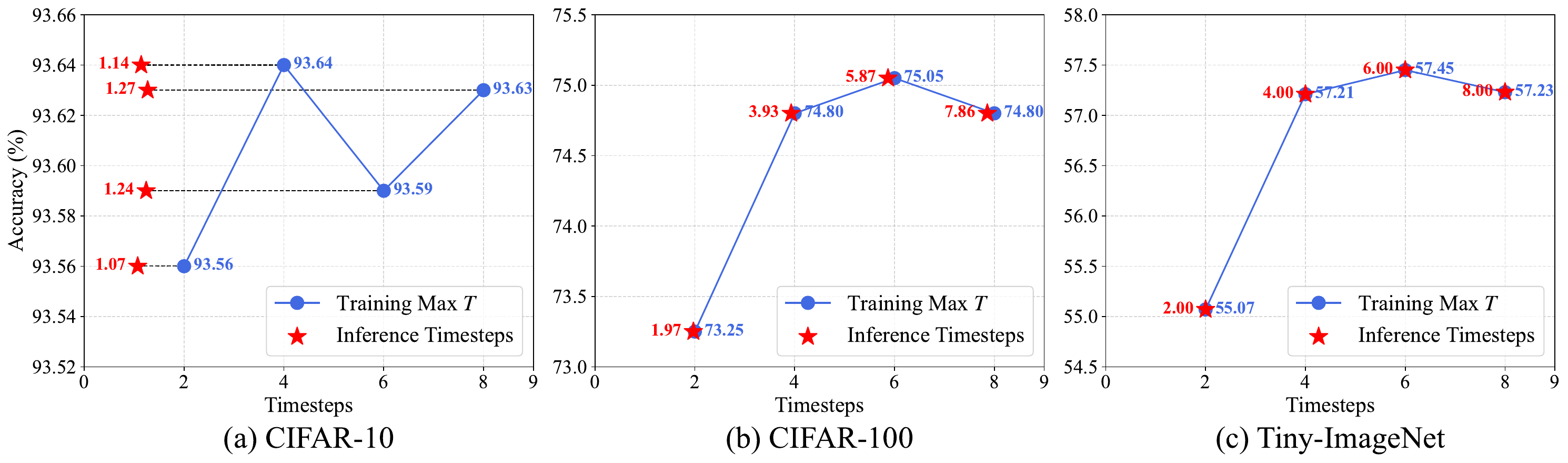}
    \caption{Evaluation of time scalability robustness. The plots show the classification accuracy (blue lines) and actual average inference timesteps on test sets (red stars) on CIFAR-10, CIFAR-100, and Tiny-ImageNet under different maximum training timesteps T}
    \label{time_scale}
\end{figure*}

In this section, we evaluate the time scalability robustness of the proposed latency-coded SNN. We train the models using the SEW-ResNet-18 architecture on the CIFAR-10, CIFAR-100, and Tiny-ImageNet datasets with maximum training timesteps $T \in \{2,4,6,8\}$. The experimental results, including both accuracy and test inference timesteps, are presented in Fig. \ref{time_scale}.

On CIFAR-100 and Tiny-ImageNet, accuracy consistently improves as T increases, reaching a peak at T=6 (75.05\% and 57.45\%, respectively) before slightly declining at T=8. This suggests that moderate temporal extension enhances feature representation, while excessive timesteps may introduce redundancy. In contrast, CIFAR-10 accuracy remains relatively stable around 93.6\% across all timesteps, indicating that our latency coding framework captures sufficient discriminative features for simpler tasks even within a minimal temporal window.

Regarding inference efficiency, the actual average inference timesteps are marked by red stars. On CIFAR-10, the network achieves early exit across multiple T, completing inference in only 1.07-1.27 timesteps. For more complex datasets, the inference timesteps scales more linearly with T to maintain precision, but still remains below the maximum T. These results demonstrate that the latency-coded SNN is robust to temporal scaling.

\subsection{Conclusion}
In this work, we presented latency coding as a practical extension of TTFS coding for deep spiking neural networks. The proposed coding scheme retains the first-spike decision principle of temporal coding while replacing floating-point firing-time communication with binary spike transmission, making it more compatible with existing neuromorphic hardware. By integrating latency-coded SNNs with BPTT, we relaxed the strict single-spike constraint in intermediate layers to improve temporal information propagation and gradient flow. In addition, the membrane-potential-assisted decoding rule enables deterministic output decisions when multiple neurons fire at the same earliest timestep.

To overcome the main optimization challenges of latency-coded SNNs, we introduced two dedicated components. The latency encoding (LE) module alleviates the information loss caused by direct intensity-to-latency mapping by first extracting salient features and then converting them into temporal spikes. The temporal adaptive decision (TAD) loss function resolves the incompatibility between latency coding and standard cross-entropy loss by dynamically weighting supervision signals according to the model's confidence, balancing the trade-off between decision speed and accuracy.

Extensive experiments demonstrate the effectiveness and scalability of the proposed framework. On CIFAR-10, CIFAR-100, Tiny-ImageNet, ImageNet, and neuromorphic datasets, latency-coded SNNs achieve competitive or superior accuracy compared with existing TTFS-based methods while reducing inference latency to only a few timesteps. Energy analysis further shows that reducing inference duration can be more important than pursuing extreme sparsity alone, since the relaxed intermediate-layer firing constraint improves trainability while the ultra-low temporal window keeps overall energy consumption low. Moreover, robustness evaluations indicate that latency coding provides stronger resistance to input perturbations than rate-coded counterparts under comparable settings. Admittedly, a marginal performance gap still exists between our approach and the latest early-exit methods of rate-coded SNNs \cite{li2023unleashing, li2023seenn}, yet our work provides a compelling alternative for energy-critical and robustness-demanding scenarios. Future research will focus on improving encoding modules, designing loss functions more closely aligned with temporal dynamics, and scaling this framework to deeper, modern architectures to fully unlock the potential of latency coding.

\bibliographystyle{IEEEtran}
\bibliography{reference}

@article{2015Batch,
  title   = {Batch Normalization: Accelerating Deep Network Training by Reducing Internal Covariate Shift},
  author  = {Ioffe, Sergey and Szegedy, Christian},
  journal = {JMLR.org},
  year    = {2015}
}

@article{1952A,
  title   = {A quantitative description of membrane current and its application to conduction and excitation in nerve.},
  author  = {Hodgkin, A. L. and Huxley, A. F.},
  journal = {Journal of Physiology},
  volume  = {117},
  year    = {1952}
}

@article{1997Networks,
  title   = {Networks of spiking neurons: the third generation of neural network models},
  author  = {Maass, Wolfgang},
  journal = {Neural Networks},
  volume  = {10},
  number  = {9},
  pages   = {1659--1671},
  year    = {1997}
}

@inproceedings{wu2019direct,
  title     = {Direct training for spiking neural networks: Faster, larger, better},
  author    = {Wu, Yujie and Deng, Lei and Li, Guoqi and Zhu, Jun and Xie, Yuan and Shi, Luping},
  booktitle = {Proceedings of the AAAI Conference on Artificial Intelligence},
  volume    = {33},
  number    = {01},
  pages     = {1311--1318},
  year      = {2019}
}

@article{deng2022temporal,
  title   = {Temporal efficient training of spiking neural network via gradient re-weighting},
  author  = {Deng, Shikuang and Li, Yuhang and Zhang, Shanghang and Gu, Shi},
  journal = {arXiv Preprint arXiv:2202.11946},
  year    = {2022}
}

@article{2021A,
  title   = {A Survey of Encoding Techniques for Signal Processing in Spiking Neural Networks},
  author  = {Auge, Daniel and Hille, Julian and Mueller, Etienne and Knoll, Alois},
  journal = {Neural Processing Letters},
  number  = {5},
  year    = {2021}
}

@article{2021Neural,
  title   = {Neural Coding in Spiking Neural Networks: A Comparative Study for Robust Neuromorphic Systems.},
  author  = {Guo, Wenzhe and Fouda, Mohammed Elneanaei and Eltawil, Ahmed and Salama, Khaled N.},
  journal = {Frontiers in Neuroscience},
  year    = {2021}
}

@inproceedings{park2020t2fsnn,
  title        = {T2FSNN: deep spiking neural networks with time-to-first-spike coding},
  author       = {Park, Seongsik and Kim, Seijoon and Na, Byunggook and Yoon, Sungroh},
  booktitle    = {2020 57th ACM/IEEE Design Automation Conference (DAC)},
  pages        = {1--6},
  year         = {2020},
  organization = {IEEE}
}

@inproceedings{han2020deep,
  title        = {Deep spiking neural network: Energy efficiency through time based coding},
  author       = {Han, Bing and Roy, Kaushik},
  booktitle    = {European Conference on Computer Vision},
  pages        = {388--404},
  year         = {2020},
  organization = {Springer}
}

@article{park2021training,
  title   = {Training energy-efficient deep spiking neural networks with time-to-first-spike coding},
  author  = {Park, Seongsik and Yoon, Sungroh},
  journal = {arXiv Preprint arXiv:2106.02568},
  year    = {2021}
}

@inproceedings{wei2023temporal,
  title     = {Temporal-coded spiking neural networks with dynamic firing threshold: Learning with event-driven backpropagation},
  author    = {Wei, Wenjie and Zhang, Malu and Qu, Hong and Belatreche, Ammar and Zhang, Jian and Chen, Hong},
  booktitle = {Proceedings of the IEEE/CVF International Conference on Computer Vision},
  pages     = {10552--10562},
  year      = {2023}
}

@article{stanojevic2024high,
  title     = {High-performance deep spiking neural networks with 0.3 spikes per neuron},
  author    = {Stanojevic, Ana and Wo{\'z}niak, Stanis{\l}aw and Bellec, Guillaume and Cherubini, Giovanni and Pantazi, Angeliki and Gerstner, Wulfram},
  journal   = {Nature Communications},
  volume    = {15},
  number    = {1},
  pages     = {6793},
  year      = {2024},
  publisher = {Nature Publishing Group UK London}
}

@article{thorpe1990spike,
  title     = {Spike arrival times: A highly efficient coding scheme for neural networks},
  author    = {Thorpe, Simon J},
  journal   = {Parallel Processing in Neural Systems},
  pages     = {91--94},
  year      = {1990},
  publisher = {Elsevier}
}

@article{optican1987temporal,
  title   = {Temporal encoding of two-dimensional patterns by single units in primate inferior temporal cortex. III. Information theoretic analysis},
  author  = {Optican, Lance M and Richmond, Barry J},
  journal = {Journal of Neurophysiology},
  volume  = {57},
  number  = {1},
  pages   = {162--178},
  year    = {1987}
}

@article{tovee1993information,
  title     = {Information encoding and the responses of single neurons in the primate temporal visual cortex},
  author    = {Tovee, Martin J and Rolls, Edmund T and Treves, Alessandro and Bellis, Raymond P},
  journal   = {Journal of Neurophysiology},
  volume    = {70},
  number    = {2},
  pages     = {640--654},
  year      = {1993},
  publisher = {American Physiological Society Bethesda, MD}
}

@article{kjaer1994decoding,
  title     = {Decoding cortical neuronal signals: network models, information estimation and spatial tuning},
  author    = {Kjaer, Troels W and Hertz, John A and Richmond, Barry J},
  journal   = {Journal of Computational Neuroscience},
  volume    = {1},
  number    = {1},
  pages     = {109--139},
  year      = {1994},
  publisher = {Springer}
}

@article{tovee1995information,
  title     = {Information encoding in short firing rate epochs by single neurons in the primate temporal visual cortex},
  author    = {Tovee, Martin J and Rolls, Edmund T},
  journal   = {Visual Cognition},
  volume    = {2},
  number    = {1},
  pages     = {35--58},
  year      = {1995},
  publisher = {Taylor \& Francis}
}

@article{hendrycks2019benchmarking,
  title   = {Benchmarking neural network robustness to common corruptions and perturbations},
  author  = {Hendrycks, Dan and Dietterich, Thomas},
  journal = {arXiv Preprint arXiv:1903.12261},
  year    = {2019}
}

@article{rathi2021diet,
  title     = {Diet-snn: A low-latency spiking neural network with direct input encoding and leakage and threshold optimization},
  author    = {Rathi, Nitin and Roy, Kaushik},
  journal   = {IEEE Transactions on Neural Networks and Learning Systems},
  volume    = {34},
  number    = {6},
  pages     = {3174--3182},
  year      = {2021},
  publisher = {IEEE}
}

@article{yao2024spike,
  title     = {Spike-based dynamic computing with asynchronous sensing-computing neuromorphic chip},
  author    = {Yao, Man and Richter, Ole and Zhao, Guangshe and Qiao, Ning and Xing, Yannan and Wang, Dingheng and Hu, Tianxiang and Fang, Wei and Demirci, Tugba and De Marchi, Michele and others},
  journal   = {Nature Communications},
  volume    = {15},
  number    = {1},
  pages     = {4464},
  year      = {2024},
  publisher = {Nature Publishing Group UK London}
}

@article{molchanov2016pruning,
  title   = {Pruning convolutional neural networks for resource efficient inference},
  author  = {Molchanov, Pavlo and Tyree, Stephen and Karras, Tero and Aila, Timo and Kautz, Jan},
  journal = {arXiv Preprint arXiv:1611.06440},
  year    = {2016}
}

@article{merolla2014million,
  title     = {A million spiking-neuron integrated circuit with a scalable communication network and interface},
  author    = {Merolla, Paul A and Arthur, John V and Alvarez-Icaza, Rodrigo and Cassidy, Andrew S and Sawada, Jun and Akopyan, Filipp and Jackson, Bryan L and Imam, Nabil and Guo, Chen and Nakamura, Yutaka and others},
  journal   = {Science},
  volume    = {345},
  number    = {6197},
  pages     = {668--673},
  year      = {2014},
  publisher = {American Association for the Advancement of Science}
}

@inproceedings{horowitz20141,
  title        = {1.1 computing's energy problem (and what we can do about it)},
  author       = {Horowitz, Mark},
  booktitle    = {2014 IEEE International Solid-State Circuits Conference Digest of Technical Papers (ISSCC)},
  pages        = {10--14},
  year         = {2014},
  organization = {IEEE}
}

@article{bu2023optimal,
  title   = {Optimal ANN-SNN conversion for high-accuracy and ultra-low-latency spiking neural networks},
  author  = {Bu, Tong and Fang, Wei and Ding, Jianhao and Dai, Penglin and Yu, Zhaofei and Huang, Tiejun},
  journal = {arXiv Preprint arXiv:2303.04347},
  year    = {2023}
}

@article{chen2023hybrid,
  title     = {A hybrid neural coding approach for pattern recognition with spiking neural networks},
  author    = {Chen, Xinyi and Yang, Qu and Wu, Jibin and Li, Haizhou and Tan, Kay Chen},
  journal   = {IEEE Transactions on Pattern Analysis and Machine Intelligence},
  volume    = {46},
  number    = {5},
  pages     = {3064--3078},
  year      = {2023},
  publisher = {IEEE}
}

@article{yin2021accurate,
  title     = {Accurate and efficient time-domain classification with adaptive spiking recurrent neural networks},
  author    = {Yin, Bojian and Corradi, Federico and Boht{\'e}, Sander M},
  journal   = {Nature Machine Intelligence},
  volume    = {3},
  number    = {10},
  pages     = {905--913},
  year      = {2021},
  publisher = {Nature Publishing Group UK London}
}

@article{panda2020toward,
  title     = {Toward scalable, efficient, and accurate deep spiking neural networks with backward residual connections, stochastic softmax, and hybridization},
  author    = {Panda, Priyadarshini and Aketi, Sai Aparna and Roy, Kaushik},
  journal   = {Frontiers in Neuroscience},
  volume    = {14},
  pages     = {653},
  year      = {2020},
  publisher = {Frontiers Media SA}
}

@article{furber2014spinnaker,
  title     = {The spinnaker project},
  author    = {Furber, Steve B and Galluppi, Francesco and Temple, Steve and Plana, Luis A},
  journal   = {Proceedings of the IEEE},
  volume    = {102},
  number    = {5},
  pages     = {652--665},
  year      = {2014},
  publisher = {IEEE}
}

@article{graves2016adaptive,
  title   = {Adaptive computation time for recurrent neural networks},
  author  = {Graves, Alex},
  journal = {arXiv Preprint arXiv:1603.08983},
  year    = {2016}
}

@article{banino2021pondernet,
  title   = {Pondernet: Learning to ponder},
  author  = {Banino, Andrea and Balaguer, Jan and Blundell, Charles},
  journal = {arXiv Preprint arXiv:2107.05407},
  year    = {2021}
}

@inproceedings{sun2025temporal,
  author    = {Sun, Qian and Lu, Chengzhuo and Chen, Wenyu and Wei, Wenjie and Wang, Jingya and Zhang, Jieyuan and Liu, Xiaoli and Ye, Yalan and Yang, Yang and Zhang, Malu},
  booktitle = {Proceedings of the 33rd ACM International Conference on Multimedia (MM)},
  title     = {Temporal-Coded Spiking {Transformer}},
  year      = {2025},
  pages     = {2616--2624},
  doi       = {10.1145/3746027.3754545}
}

@article{li2017cifar10,
  title     = {Cifar10-dvs: an event-stream dataset for object classification},
  author    = {Li, Hongmin and Liu, Hanchao and Ji, Xiangyang and Li, Guoqi and Shi, Luping},
  journal   = {Frontiers in Neuroscience},
  volume    = {11},
  pages     = {244131},
  year      = {2017},
  publisher = {Frontiers}
}

@article{simonyan2014very,
  title   = {Very deep convolutional networks for large-scale image recognition},
  author  = {Simonyan, Karen and Zisserman, Andrew},
  journal = {arXiv Preprint arXiv:1409.1556},
  year    = {2014}
}

@article{fang2021deep,
  title   = {Deep residual learning in spiking neural networks},
  author  = {Fang, Wei and Yu, Zhaofei and Chen, Yanqi and Huang, Tiejun and Masquelier, Timoth{\'e}e and Tian, Yonghong},
  journal = {Advances in Neural Information Processing Systems},
  volume  = {34},
  pages   = {21056--21069},
  year    = {2021}
}

@article{loshchilov2017decoupled,
  title   = {Decoupled weight decay regularization},
  author  = {Loshchilov, Ilya and Hutter, Frank},
  journal = {arXiv Preprint arXiv:1711.05101},
  year    = {2017}
}

@inproceedings{cubuk2019autoaugment,
  title     = {Autoaugment: Learning augmentation strategies from data},
  author    = {Cubuk, Ekin D and Zoph, Barret and Mane, Dandelion and Vasudevan, Vijay and Le, Quoc V},
  booktitle = {Proceedings of the IEEE/CVF Conference on Computer Vision and Pattern Recognition},
  pages     = {113--123},
  year      = {2019}
}

@inproceedings{cubuk2020randaugment,
  title     = {Randaugment: Practical automated data augmentation with a reduced search space},
  author    = {Cubuk, Ekin D and Zoph, Barret and Shlens, Jonathon and Le, Quoc V},
  booktitle = {Proceedings of the IEEE/CVF Conference on Computer Vision and Pattern Recognition Workshops},
  pages     = {702--703},
  year      = {2020}
}

@article{le2015tiny,
  title   = {Tiny imagenet visual recognition challenge},
  author  = {Le, Yann and Yang, Xuan},
  journal = {CS 231N},
  volume  = {7},
  number  = {7},
  pages   = {3},
  year    = {2015}
}

@inproceedings{deng2009imagenet,
  title        = {Imagenet: A large-scale hierarchical image database},
  author       = {Deng, Jia and Dong, Wei and Socher, Richard and Li, Li-Jia and Li, Kai and Fei-Fei, Li},
  booktitle    = {2009 IEEE Conference on Computer Vision and Pattern Recognition},
  pages        = {248--255},
  year         = {2009},
  organization = {IEEE}
}

@article{ILSVRC15,
  author  = {Russakovsky, Olga and Deng, Jia and Su, Hao and Krause, Jonathan and Satheesh, Sanjeev and Ma, Sean and Huang, Zhiheng and Karpathy, Andrej and Khosla, Aditya and Bernstein, Michael and Berg, Alexander C. and Fei-Fei, Li},
  title   = {ImageNet Large Scale Visual Recognition Challenge},
  year    = {2015},
  journal = {International Journal of Computer Vision (IJCV)},
  doi     = {10.1007/s11263-015-0816-y},
  volume  = {115},
  number  = {3},
  pages   = {211--252}
}

@inproceedings{li2022neuromorphic,
  title        = {Neuromorphic data augmentation for training spiking neural networks},
  author       = {Li, Yuhang and Kim, Youngeun and Park, Hyoungseob and Geller, Tamar and Panda, Priyadarshini},
  booktitle    = {European Conference on Computer Vision},
  pages        = {631--649},
  year         = {2022},
  organization = {Springer}
}

@inproceedings{li2023unleashing,
  title     = {Unleashing the potential of spiking neural networks with dynamic confidence},
  author    = {Li, Chen and Jones, Edward G and Furber, Steve},
  booktitle = {Proceedings of the IEEE/CVF International Conference on Computer Vision},
  pages     = {13350--13360},
  year      = {2023}
}

@article{li2023seenn,
  title   = {Seenn: Towards temporal spiking early exit neural networks},
  author  = {Li, Yuhang and Geller, Tamar and Kim, Youngeun and Panda, Priyadarshini},
  journal = {Advances in Neural Information Processing Systems},
  volume  = {36},
  pages   = {63327--63342},
  year    = {2023}
}

@article{perez2013mapping,
  title     = {Mapping from frame-driven to frame-free event-driven vision systems by low-rate rate coding and coincidence processing--application to feedforward ConvNets},
  author    = {P{\'e}rez-Carrasco, Jos{\'e} Antonio and Zhao, Bo and Serrano, Carmen and Acha, Begona and Serrano-Gotarredona, Teresa and Chen, Shouchun and Linares-Barranco, Bernab{\'e}},
  journal   = {IEEE Transactions on Pattern Analysis and Machine Intelligence},
  volume    = {35},
  number    = {11},
  pages     = {2706--2719},
  year      = {2013},
  publisher = {IEEE}
}

@inproceedings{han2020rmp,
  title     = {Rmp-snn: Residual membrane potential neuron for enabling deeper high-accuracy and low-latency spiking neural network},
  author    = {Han, Bing and Srinivasan, Gopalakrishnan and Roy, Kaushik},
  booktitle = {Proceedings of the IEEE/CVF Conference on Computer Vision and Pattern Recognition},
  pages     = {13558--13567},
  year      = {2020}
}

@article{neftci2019surrogate,
  title     = {Surrogate gradient learning in spiking neural networks: Bringing the power of gradient-based optimization to spiking neural networks},
  author    = {Neftci, Emre O and Mostafa, Hesham and Zenke, Friedemann},
  journal   = {IEEE Signal Processing Magazine},
  volume    = {36},
  number    = {6},
  pages     = {51--63},
  year      = {2019},
  publisher = {IEEE}
}

@article{wu2018spatio,
  title     = {Spatio-temporal backpropagation for training high-performance spiking neural networks},
  author    = {Wu, Yujie and Deng, Lei and Li, Guoqi and Zhu, Jun and Shi, Luping},
  journal   = {Frontiers in Neuroscience},
  volume    = {12},
  pages     = {331},
  year      = {2018},
  publisher = {Frontiers Media SA}
}

@article{qin2023attention,
  title     = {Attention-based deep spiking neural networks for temporal credit assignment problems},
  author    = {Qin, Lang and Wang, Ziming and Yan, Rui and Tang, Huajin},
  journal   = {IEEE Transactions on Neural Networks and Learning Systems},
  volume    = {35},
  number    = {8},
  pages     = {10301--10311},
  year      = {2023},
  publisher = {IEEE}
}

@article{ding2021optimal,
  title   = {Optimal ANN-SNN conversion for fast and accurate inference in deep spiking neural networks},
  author  = {Ding, Jianhao and Yu, Zhaofei and Tian, Yonghong and Huang, Tiejun},
  journal = {arXiv Preprint arXiv:2105.11654},
  year    = {2021}
}

@inproceedings{bing2018end,
  title        = {End to end learning of spiking neural network based on r-stdp for a lane keeping vehicle},
  author       = {Bing, Zhenshan and Meschede, Claus and Huang, Kai and Chen, Guang and Rohrbein, Florian and Akl, Mahmoud and Knoll, Alois},
  booktitle    = {2018 IEEE International Conference on Robotics and Automation (ICRA)},
  pages        = {4725--4732},
  year         = {2018},
  organization = {IEEE}
}

@article{bing2020indirect,
  title     = {Indirect and direct training of spiking neural networks for end-to-end control of a lane-keeping vehicle},
  author    = {Bing, Zhenshan and Meschede, Claus and Chen, Guang and Knoll, Alois and Huang, Kai},
  journal   = {Neural Networks},
  volume    = {121},
  pages     = {21--36},
  year      = {2020},
  publisher = {Elsevier}
}

@article{wu2021progressive,
  title     = {Progressive tandem learning for pattern recognition with deep spiking neural networks},
  author    = {Wu, Jibin and Xu, Chenglin and Han, Xiao and Zhou, Daquan and Zhang, Malu and Li, Haizhou and Tan, Kay Chen},
  journal   = {IEEE Transactions on Pattern Analysis and Machine Intelligence},
  volume    = {44},
  number    = {11},
  pages     = {7824--7840},
  year      = {2021},
  publisher = {IEEE}
}

@article{wu2021tandem,
  title     = {A tandem learning rule for effective training and rapid inference of deep spiking neural networks},
  author    = {Wu, Jibin and Chua, Yansong and Zhang, Malu and Li, Guoqi and Li, Haizhou and Tan, Kay Chen},
  journal   = {IEEE Transactions on Neural Networks and Learning Systems},
  volume    = {34},
  number    = {1},
  pages     = {446--460},
  year      = {2021},
  publisher = {IEEE}
}

@article{kheradpisheh2018stdp,
  title     = {STDP-based spiking deep convolutional neural networks for object recognition},
  author    = {Kheradpisheh, Saeed Reza and Ganjtabesh, Mohammad and Thorpe, Simon J and Masquelier, Timoth{\'e}e},
  journal   = {Neural Networks},
  volume    = {99},
  pages     = {56--67},
  year      = {2018},
  publisher = {Elsevier}
}

@article{mostafa2017supervised,
  title     = {Supervised learning based on temporal coding in spiking neural networks},
  author    = {Mostafa, Hesham},
  journal   = {IEEE Transactions on Neural Networks and Learning Systems},
  volume    = {29},
  number    = {7},
  pages     = {3227--3235},
  year      = {2017},
  publisher = {IEEE}
}

@article{yu2013precise,
  title     = {Precise-spike-driven synaptic plasticity: Learning hetero-association of spatiotemporal spike patterns},
  author    = {Yu, Qiang and Tang, Huajin and Tan, Kay Chen and Li, Haizhou},
  journal   = {PLOS ONE},
  volume    = {8},
  number    = {11},
  pages     = {e78318},
  year      = {2013},
  publisher = {Public Library of Science San Francisco, USA}
}

@article{zhang2021rectified,
  title     = {Rectified linear postsynaptic potential function for backpropagation in deep spiking neural networks},
  author    = {Zhang, Malu and Wang, Jiadong and Wu, Jibin and Belatreche, Ammar and Amornpaisannon, Burin and Zhang, Zhixuan and Miriyala, Venkata Pavan Kumar and Qu, Hong and Chua, Yansong and Carlson, Trevor E and others},
  journal   = {IEEE Transactions on Neural Networks and Learning Systems},
  volume    = {33},
  number    = {5},
  pages     = {1947--1958},
  year      = {2021},
  publisher = {IEEE}
}

@article{ding2025assisting,
  title     = {Assisting training of deep spiking neural networks with parameter initialization},
  author    = {Ding, Jianhao and Zhang, Jiyuan and Huang, Tiejun and Liu, Jian K and Yu, Zhaofei},
  journal   = {IEEE Transactions on Neural Networks and Learning Systems},
  year      = {2025},
  publisher = {IEEE}
}

@article{liu2026fpf,
  title     = {FPF-SNNs: Floating-Point-Free Spiking Neural Networks},
  author    = {Liu, Hanwen and Cai, Chiyu and Shi, Kexin and Wei, Wenjie and Zhang, Jieyuan and Chen, Wenyu and Zhang, Malu and Yang, Yang},
  journal   = {IEEE Transactions on Emerging Topics in Computational Intelligence},
  year      = {2026},
  publisher = {IEEE}
}

@article{tang2024neuromorphic,
  title     = {Neuromorphic auditory perception by neural spiketrum},
  author    = {Tang, Huajin and Gu, Pengjie and Wijekoon, Jayawan and Alsakkal, MHD Anas and Wang, Ziming and Shen, Jiangrong and Yan, Rui and Pan, Gang},
  journal   = {IEEE Transactions on Emerging Topics in Computational Intelligence},
  volume    = {9},
  number    = {1},
  pages     = {292--303},
  year      = {2024},
  publisher = {IEEE}
}

@article{jiang2024deep,
  title     = {Deep spiking neural networks driven by adaptive interval membrane potential for temporal credit assignment problem},
  author    = {Jiang, Jiaqiang and Ding, Haohui and Wang, Haixia and Yan, Rui},
  journal   = {IEEE Transactions on Emerging Topics in Computational Intelligence},
  volume    = {9},
  number    = {1},
  pages     = {717--728},
  year      = {2024},
  publisher = {IEEE}
}

@article{shen2025exploiting,
  title     = {Exploiting high performance spiking neural networks with efficient spiking patterns},
  author    = {Shen, Guobin and Zhao, Dongcheng and Zeng, Yi},
  journal   = {IEEE Transactions on Emerging Topics in Computational Intelligence},
  year      = {2025},
  publisher = {IEEE}
}

\end{document}